\documentclass[lettersize,journal]{IEEEtran}

\usepackage[utf8]{inputenc}
\usepackage[english]{babel}
\usepackage{algorithm}
\usepackage[noend]{algorithmic}
\usepackage{tcolorbox}
\usepackage{tikz}
\usepackage{pgfplots}
\usepackage{pgfplotstable}
\usepackage{multirow}
\usepackage{soul}
\usepackage{lipsum}     
\usepackage{pifont}
\usepackage{xargs} 
\usepackage{siunitx}
\usepackage{graphicx}
\usepackage{color}
\usepackage{epstopdf}
\usepackage{url}
\usepackage{acronym}
\usepackage[absolute,overlay]{textpos}
\usepackage{paralist}
\usepackage[normalem]{ulem}
\usepackage[caption=false]{subfig}
\usepackage{xcolor}
\definecolor{grey}{rgb}{0.5,0.5,0.5}
\usepackage{orcidlink}
\usepackage{hyperref}
\usepackage{amsmath}

\usepackage{amsfonts}
\usepackage{tcolorbox}
\usepackage{etoolbox}
\usepackage{cite}
\usepackage{booktabs}
\usepackage[absolute,overlay]{textpos}

\patchcmd{\thebibliography}{\normalsize}{\footnotesize}{}{}

\usepackage{listings}
\pgfplotsset{compat=1.12}
\definecolor{Yellow}{rgb}{1,0.9,0.7}
\definecolor{Pink}{rgb}{1,0.85,0.85}
\definecolor{AntiqueWhite}{rgb}{0.9,0.9,0.9}
\definecolor{darkgreen}{RGB}{0,100,0}

\newcommand{\fix}[1]%
{
\noindent
\fboxsep=2mm\fcolorbox{black}{AntiqueWhite}{\parbox{0.95\columnwidth}
{\textbf{AK: } #1}
}
}

\newcommand{\NOTE}[1]%
{
\noindent
\fboxsep=2mm\fcolorbox{black}{AntiqueWhite}{\parbox{0.95\columnwidth}
{\textbf{NOTE: } #1}
}
}

\hypersetup{
    colorlinks=true,
    citecolor=blue,
    linkcolor=blue,
    filecolor=magenta,      
    urlcolor=black,
    pdftitle={TAPAS: Throughput-adaptive Perception for Autonomous Systems},
    pdfpagemode=FullScreen,
    }

\title{TAPAS: Throughput-adaptive Perception for Autonomous Systems}

\author{
Aman Vyas
, Vasista Kodumagulla, 
Zain Taufique, 
Pasi Liljeberg, 
Anil Kanduri 

Department of Computing, University of Turku, Finland

Email: \{amvyas, vakodu, zatauf, pakrli, spakan\}@utu.fi
}

\begin{document}
\maketitle

\begin{textblock*}{3cm}(20cm,8cm) 
\rotatebox{90}{\textit{Accepted at ACM/IEEE International Conference on Codesign of Embedded Systems (ESWEEK-CODES), 2026}}
\end{textblock*}

\begin{abstract}

Autonomous systems rely on a perception module to navigate through dynamic environments. In real-world scenarios, the perception module's throughput requirements vary at runtime due to changes in scene complexity. 
However, existing perception strategies assume a fixed FPS and static model-to-cluster mapping, resulting in either over/under provision of throughput requirements or unnecessary energy consumption across diverse scenes. Addressing this challenge requires tightly coupled \textit{scene complexity awareness} to estimate an appropriate FPS target and \textit{dynamic model-to-cluster mapping} to deliver the required throughput at minimum energy. We propose a throughput-adaptive perception strategy for mobile/edge platforms, enabling intelligent runtime resource allocation based on varying FPS targets. We use Reinforcement Learning (RL) with RRM (Reward Reasoning Model) and a GRU (Gated Recurrent Unit) agent to orchestrate perception tasks across heterogeneous mobile/edge platforms. We evaluate TAPAS on Jetson Orin NX across KITTI and unseen nuScenes. On the \textit{KITTI} dataset's test sequences, TAPAS achieves 93-100\% throughput met rate while saving energy by 76\%. On the unseen \textit{nuScenes} dataset, TAPAS maintains 97\% throughput met rate with 64\% lower energy compared to \textit{SOTA} approaches, proving its robustness.


\end{abstract}







\acrodef{as}[AS]{Autonomous systems}
\acrodef{mas}[mAS]{Mobile Autonomous systems}
\acrodef{dla}[NVDLA]{NVIDIA Deep Learning Accelerator}
\acrodef{cpu}[CPU]{Central Processing Unit}
\acrodef{gpu}[GPU]{Graphics Processing Unit}
\acrodef{npu}[NPU]{Neural Processing Unit}
\acrodef{dnn}[DNN]{Deep Neural Network}
\acrodef{hmp}[HMP]{Heterogeneous Multi-core Processing}
\acrodef{fps}[FPS]{Frames Per Second}
\acrodef{DVFS}[DVFS]{Dynamic Voltage/Frequency Scaling}
\acrodef{rl}[RL]{Reinforcement Learning}
\acrodef{gru}[GRU]{Gated Recurrent Unit}
\acrodef{rrm}[RRM]{Reward Reasoning Model}
\acrodef{LLM}[LLM]{Large Language Model}
\acrodef{ppo}[PPO]{Proximal Policy Optimization}
\acrodef{grm}[GRM]{Generative reward modeling}
\acrodef{sota}[SOTA]{State-of-the-art}

\acused{cpu}
\acused{gpu}




\section{introduction} \label{sec.intro}

\ac{as} such as self-driving vehicles, drones, robots, etc., rely on a perception stack for navigation, planning, and control~\cite{inspection_ral}. Modern perception pipelines include deep learning based vision tasks for object detection, semantic segmentation, visual odometry, and obstacle avoidance \cite{tsformer_vo,vit}. Throughput of the perception module (typically expressed in frames-per-second (FPS)~\cite{roboshape_isca,zhuyi_dac}) is crucial for efficient navigation of \ac{as}~\cite{How_fast_is_too_fast}. The throughput of the perception module depends on the complexity of the scene in a given frame~\cite{zhuyi_dac}. Complex heterogeneous scenes (e.g., dense urban environments) require processing at a higher FPS for efficient navigation, while processing at a lower FPS suffices for sparse homogeneous scenes\cite{zhuyi_dac, How_fast_is_too_fast, percetion_aware_iccd, roborun_dac,context_date}.

In real-world navigation, \ac{as} encounter a mixture of both simple and complex scenes, creating variable FPS requirements over time. Yet, existing perception modules operate at a single FPS target that is fixed at design time, irrespective of the dynamic scene complexity~\cite{How_fast_is_too_fast, roboshape_isca}. Setting a conservatively higher FPS target offers robust navigation in dense scenes. However, this approach delivers unnecessarily higher FPS in sparse environments~\cite{How_fast_is_too_fast,zhuyi_dac}, resulting in significant computational and energy wastage that is particularly detrimental to battery-powered mobile autonomous systems. Thus, \ac{as} operating in unstructured and dynamic environments requires a perception module that can adapt to variable FPS targets for energy efficiency.

\begin{figure}
    \centering
    \includegraphics[width=0.99\linewidth]{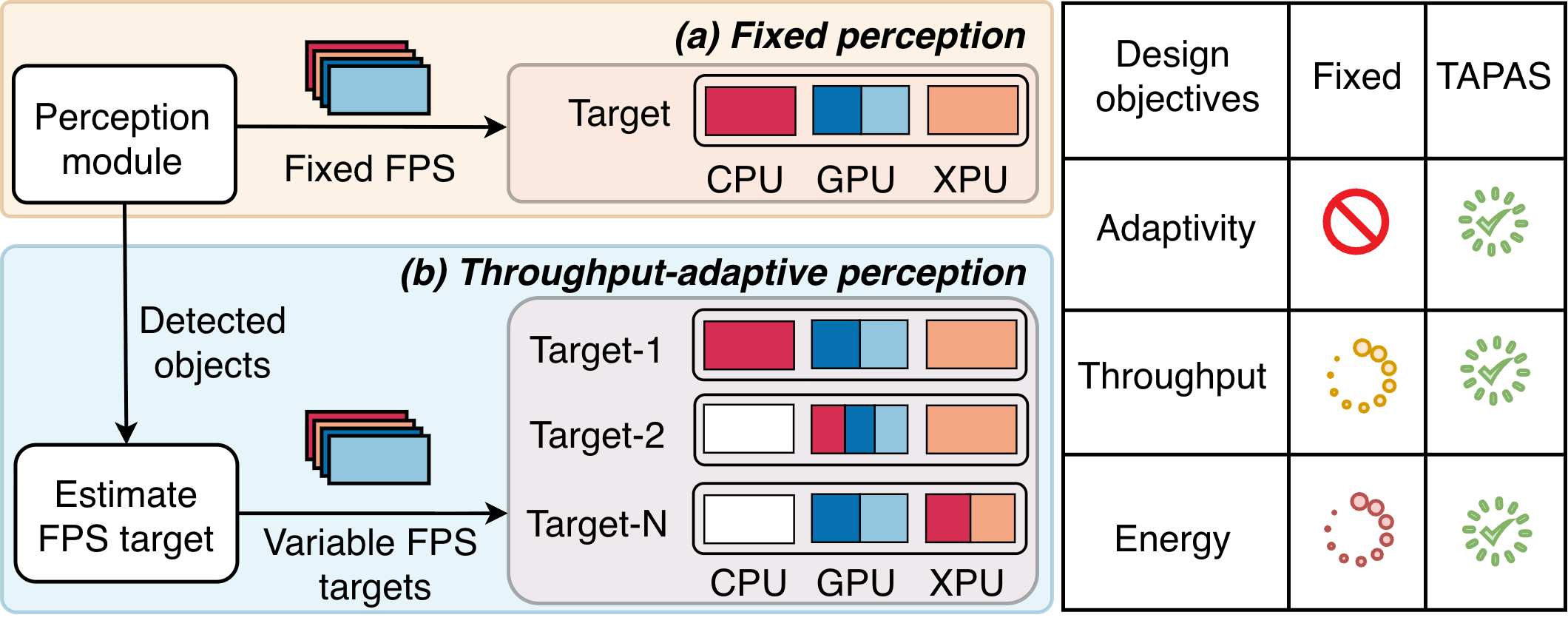}
    \caption{Overview of static and adaptive perception modules: SOTA approaches (top) employ fixed FPS and static mapping; TAPAS (bottom region) introduces estimating FPS targets and runtime mapping.}
    \label{fig:intro}
    \vspace{-15pt}
\end{figure}

\ac{as} are increasingly employing \acp{hmp} with asymmetric CPUs, GPUs, and custom deep learning accelerators \cite{dla_manual}. Core-level heterogeneity offers flexibility in running the perception pipeline at different FPS targets and energy budgets. Despite using accelerator-rich hardware, existing perception pipelines employ static model-to-cluster mapping approaches ~\cite{ee_iros,ee_tro}, restricting their throughput adaptivity. State-of-the-art inference optimization techniques for \acp{hmp}~\cite{band_npu,omniboost,Axonn_dla} cannot be directly repurposed for the perception pipeline, since they: (i) lack high-level domain semantic awareness (e.g., scene complexity), (ii) are confined to a single class of vision tasks (such as object detection), as opposed to concurrently running multiple classes of vision modules in a modern perception pipeline, and (iii) are not designed for multi-objective throughput adaptivity.

Adapting to varying throughput targets requires tightly coupled \textit{scene complexity awareness} -- to estimate an appropriate FPS target, and \textit{dynamic model-to-cluster mapping} -- to deliver the scene complexity-aware throughput target with the lowest energy consumption. Figure~\ref{fig:intro} (top) illustrates a representation of existing perception modules with a fixed FPS target and static model-to-cluster mapping. Without scene awareness and run-time adaptability, such strategies can provide only a predetermined throughput across diverse scenes, draining compute and energy resources. Figure~\ref{fig:intro} (bottom) demonstrates \textit{throughput-adaptive perception}, where variable FPS targets (Target 1, 2,..N per scene 1, 2,..N) are estimated through scene awareness (Section \ref{sec.bac_spatial_entpy}~\cite{spatial_entropy}). The scene-derived FPS target is parsed to update the model-to-cluster mapping at run-time, ensuring that only the required FPS target is delivered, significantly minimizing energy consumption. 

\textit{Our goal} is to design a throughput-adaptive perception framework for \ac{as} running on heterogeneous mobile/edge platforms. We propose \textit{TAPAS}, a framework that estimates FPS targets based on scene complexity and dynamically configures compute resources to meet these targets while minimizing energy consumption. \textit{TAPAS} uses \textit{spatial entropy} computed from perception pipeline outputs using Shannon's entropy (Section~\ref{sec.spatial_entropy}) to estimate scene-specific FPS targets. This lightweight mechanism quantifies scene complexity by directly mapping entropy levels to FPS targets. Configuring compute resources to meet variable FPS targets presents a complex design space exploration challenge due to application diversity (multiple DNN and transformer models in the perception pipeline) and hardware diversity (different energy-latency characteristics and variable deep learning operator support across clusters). Navigating such complex design space through exhaustive search is infeasible, heuristic methods lack adaptivity beyond design-time assumptions, and supervised learning cannot generalize beyond labeled data. In this context, \ac{rl} enables adaptation to variable FPS demands through system-level reward signals without labeled training data, by learning policies that maximize long-term returns across diverse \ac{hmp} configurations. This enables TAPAS to make zero-shot predictions across unseen scenes and perception pipelines. We adopt \ac{ppo}~\cite{PPO} \ac{rl}, which provides stable policy updates via clipped surrogate objectives and sample-efficient learning (Section~\ref{sec.rl}) compared to heuristic-based methods. Despite \ac{rl}'s advantages, two challenges remain open. First, scene complexity exhibits temporal dependencies that memory-less agents~\cite{omniboost, tango} cannot exploit. We address this using a \ac{gru} agent (Section~\ref{sec.gru}) that leverages variable FPS, entropy trajectories, and workload variations to perform model-to-cluster mapping. Second, joint throughput-energy optimization involves conflicting objectives that simple weighted-sum or heuristic reward functions cannot resolve~\cite{tango,omniboost}. We address this through \ac{rrm}~\cite{reasoning_reward_model} (Section~\ref{sec.rrm}) that provides structured reasoning to balance throughput met rate and energy efficiency under variable FPS demands. Our contributions are:

\begin{figure}
    \centering
    \includegraphics[width=0.99\linewidth]{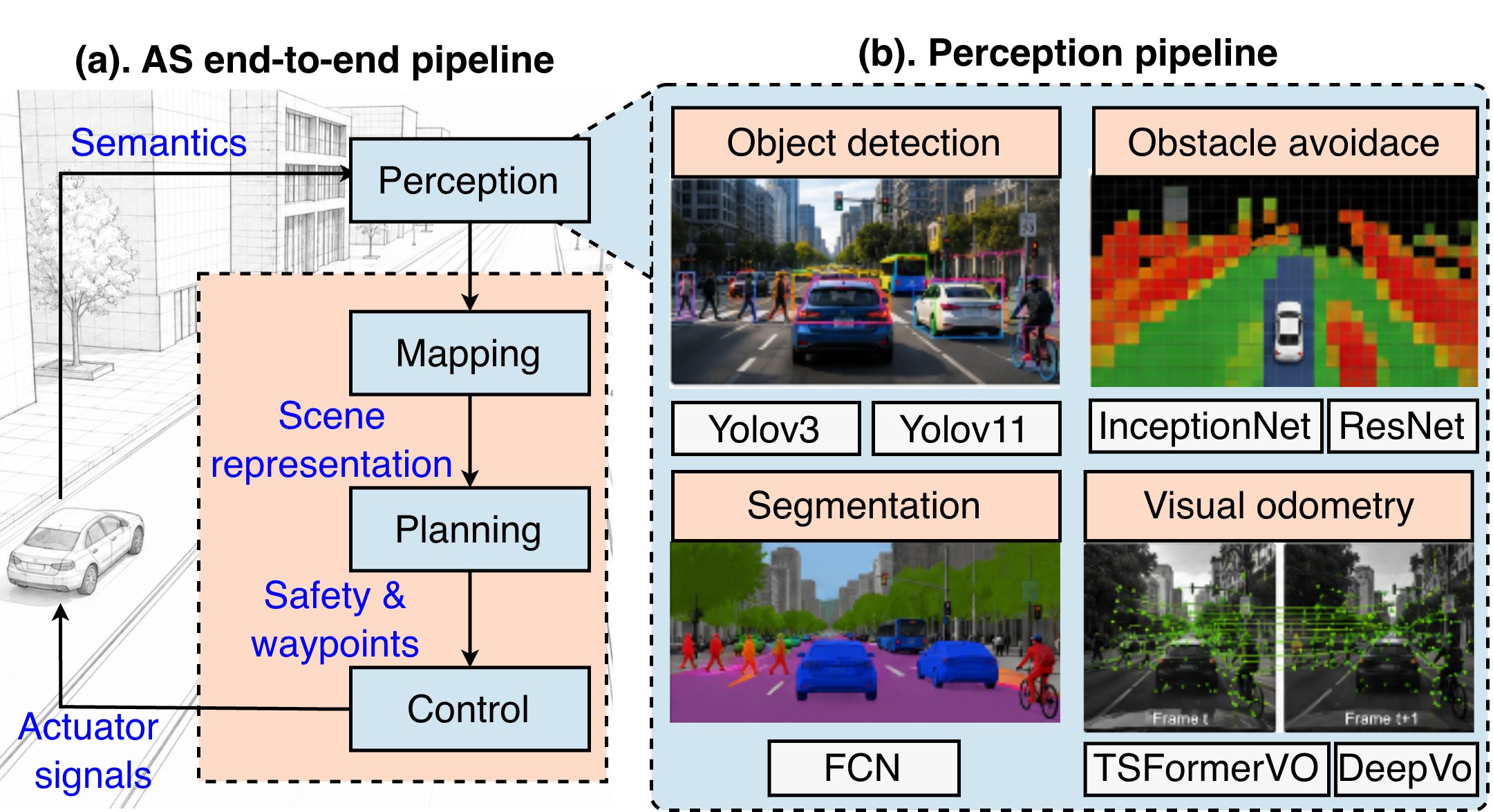}
    \caption{End-to-end and perception pipeline for AS.}
    \vspace{-12pt}
    \label{fig:back_perception}
\end{figure}

\begin{itemize}
  
    \item TAPAS framework for orchestrating throughput-adaptive perception on mobile \acp{hmp}, by coupling scene-awareness with system-level resource allocation.
    
   \item Quantifying scene dynamicity with spatial entropy to estimate run-time variable throughput targets for throughput-adaptivity. 
   
    \item Designing a \ac{gru}-based \ac{rl} agent with \ac{rrm} for capturing temporal context while balancing energy-throughput objectives, to determine model-to-cluster mapping under varying throughput targets.
    
    \item Deployment and evaluation of TAPAS on Jetson Orin NX platform, demonstrating up to 93–100\% throughput-met rate with up to 76\% energy savings on KITTI sequences, and generalizes to unseen nuScenes data with 97\% throughput-met rate and 64\% lower energy.   
\end{itemize}

\begin{figure}
    \centering
    \includegraphics[width=0.99\linewidth]{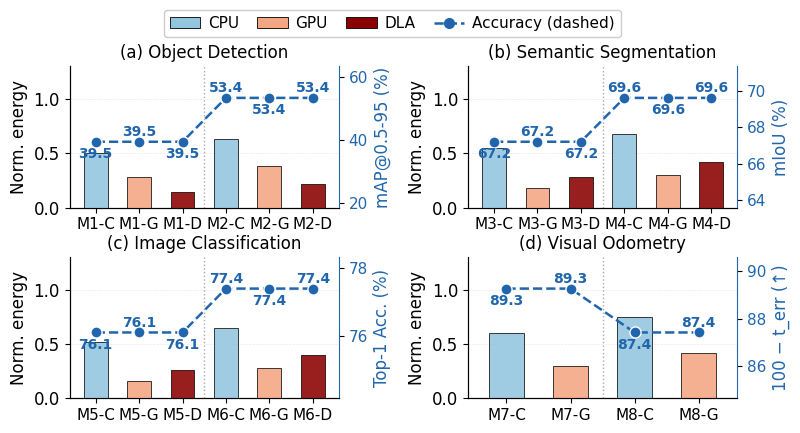}
    \caption{Impact of hardware vs. algorithmic knobs on energy and accuracy of perception pipeline}
    \label{fig:back_0}
    \vspace{-12pt}
\end{figure}

\section{Background and Motivation} \label{sec.background}

\subsection{Background}

\subsubsection{Autonomous systems pipeline and perception module overview}
As presented in figure \ref{fig:back_perception}(a), \ac{as} contains functionally independent yet semantically coupled modules: Perception, Mapping, Planning, and Control. At the task level, each module operates independently and can be designed, profiled, and optimized in isolation. However, at the functional level, these modules create a strict dependency chain: mapping consumes perception outputs to build scene representations, planning relies on the map to reason about waypoints and safety, and control translates planned trajectories into real-time actuator signals. As highlighted in figure \ref{fig:back_perception}(b), perception module encompasses four concurrent vision tasks: object detection, obstacle avoidance, semantic segmentation, and visual odometry. Each perception task is implemented by a dedicated DNN and/or transformer. The perception module constitutes the most compute and energy-intensive stage of the \ac{as} stack. The sole responsibility of perception is to extract scene semantics from sensor data, while safety and motion planning are handled by downstream planning modules. This separation decouples perception from safety-critical reasoning, enabling enhanced scene understanding while allowing downstream modules to independently manage navigation and safety decisions.

\subsubsection{Significance of run-time system for adaptive perception}
With the increasing diversity of perception workload and the increasing heterogeneity of edge platforms, efficiently deploying perception pipelines on \ac{hmp} has become a critical challenge. \textit{TAPAS} addresses this through a portable, cross-platform middleware that provides runtime hardware-level adaptation, avoiding the need for application-specific model optimization. Consequently, it generalizes across diverse shallow and deep perception pipelines for \ac{as}. One approach is to tune the perception pipeline through algorithmic knobs, such as model approximation, quantization, and pruning, which typically trade accuracy for improvements in throughput and energy efficiency. Figure \ref{fig:back_0} depicts the resulting energy–accuracy trade-off space enabled by algorithmic and hardware adaptation mechanisms. We characterize the adaptivity on \ac{hmp} platform (Jetson Orin NX) comprising CPU (C), GPU (G), and DLA (D), from both algorithmic and hardware perspectives. Our study considers perception models, including YOLOv11-n/l (M1–M2), FCN-ResNet50/101 (M3–M4), ResNet50/101 (M5–M6), and TSformerVO-1/2 (M7–M8), executed on the Jetson Orin NX platform. We evaluate object detection using mAP, semantic segmentation using mIoU, image classification using Top-1 accuracy, and visual odometry using translation error. Typically, algorithmic knobs achieve adaptivity by trading accuracy to meet varying FPS targets. An algorithmic knob, such as model approximation, performs model selection (e.g., (M-1 $\leftrightarrow$ M-2), (M-3 $\leftrightarrow$ M-4), (M-5 $\leftrightarrow$ M-6), and (M-7 $\leftrightarrow$ M-8)) to meet varying FPS targets.  In the case of model approximation, perception quality is degraded by switching between models of different accuracies to achieve the desired FPS target. While this improves adaptivity, it can compromise the safety of the \ac{as} in dynamic environments due to reduced perception quality. Furthermore, since \textit{TAPAS} relies on object-detection outputs to estimate spatial entropy and derive FPS targets, runtime model switching can alter detection quality and consequently affect the fidelity of entropy estimation itself. In contrast, hardware knobs such as cluster selection and task migration treat model accuracy as a fixed system parameter. Different clusters exhibit distinct performance–energy characteristics, and for variable FPS targets this heterogeneity can be leveraged to improve efficiency without degrading perception quality. Specifically, tasks can be migrated based on performance-to-cluster affinity by exploiting the diversity in compute and energy profiles across clusters. This approach provides adaptivity through throughput-energy trade-offs without degrading perception quality. Therefore, in \textit{TAPAS}, we leverage hardware knobs (cluster mapping and task migration) to achieve adaptive and energy-efficient operation while preserving the accuracy of the throughput estimator.

\begin{figure}
    \centering
    \includegraphics[width=0.99\linewidth]{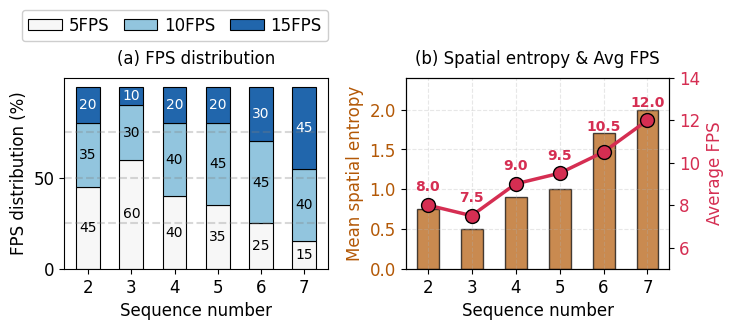}
    \caption{(a) Distribution of FPS targets across KITTI sequences. (b) Mean spatial entropy-throughput correlation.}
    \vspace{-18.5pt}
    \label{fig:rpp}
\end{figure}

\subsection{Motivation}

\subsubsection{Impact of spatial entropy on throughput estimation}
\label{sec.bac_spatial_entpy}

Existing perception strategies lack domain-semantic awareness \cite{adavp,context_date,confined_inspection}, limiting them from considering environment-driven, run-time variable FPS targets. Spatial entropy is widely used to quantify the dynamism in a sequence of images \cite{Shannon_spatial_entropy}. We propose using \textit{spatial entropy} (section \ref{sec.spatial_entropy}) to estimate variable FPS targets based on scene complexity, thereby enabling throughput adaptivity. We demonstrate the correlation between spatial entropy and FPS targets using test sequences from the \textit{KITTI} dataset. Typical perception module in \ac{as} operates at an average of 10 Hz (10 FPS) \cite{roboshape_isca}. For simplicity of demonstration, we mapped spatial entropy values of the KITTI sequences into 3 FPS levels (5, 10, and 15), with 10 FPS as the baseline requirement. Figure \ref{fig:rpp}(a) shows the distribution of estimated FPS target levels for different sequences. It should be noted that spatial entropy varies significantly across the test sequences, as reflected in the FPS requirements. Figure \ref{fig:rpp}(b) shows the mean spatial entropy of test sequences 2-7. For instance, sequence 7 has the highest spatial entropy with a 12 FPS target (average over the entire sequence), whereas sequence 3 has the lowest spatial entropy with 7.5 FPS target (on average). This variation in FPS targets (1.6x) motivates the need for throughput-adaptive resource allocation, driven by scene complexity.

\subsubsection{Significance of throughput-adaptive perception} \label{sec.sig_ada_th}

\begin{figure}
    \centering
    \includegraphics[width=0.99\linewidth]{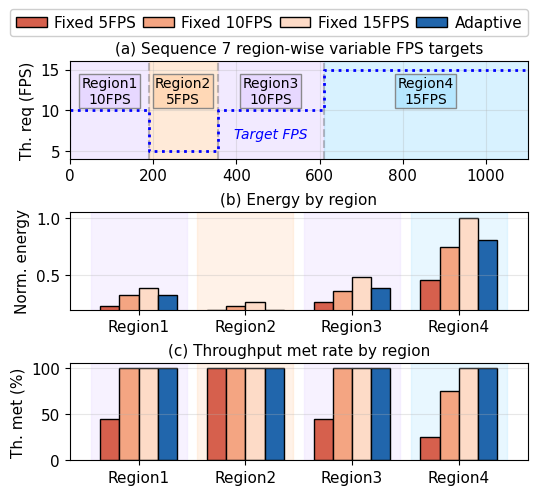}
    \caption{(a). Region-wise FPS targets in KITTI Sequence 7, (b). Normalized energy, and (c). Throughput target met rate with fixed and adaptive perception}
    \label{fig:back_3}
    \vspace{-15pt}
\end{figure}
 
We present the utility of throughput-adaptive perception over an exemplar dynamic test sequence 7 from KITTI dataset. For perception, we consider SOTA perception pipeline with FCN, Inception, YOLOv3-tiny, and TSFormerVO models. We use Jetson Orin NX (hexa-core ARM Cortex-A78 CPU, Ampere GPU with 1024 CUDA cores, and NVDLA) as the target edge AI platform. Figure~\ref{fig:back_3}(a) shows KITTI sequence 7, which is composed of four regions with variable FPS targets (5, 10, and 15 FPS), estimated using spatial entropy. We demonstrate the disparity between fixed and adaptive perception across all the regions. In Figure~\ref{fig:back_3}(a), variable FPS targets are shown in a blue dotted line. Fixed 5 FPS under-provisions for Regions 1, 3, and 4; Fixed 10 FPS target under-provisions for region 4, while over-provisioning for region 2, and Fixed 15 FPS meets all FPS targets but over-provisions for low-complexity scenes (region 1-3). This highlights the fundamental gap between fixed and scene-aware dynamic FPS targets. In contrast, adaptive perception uses a variable FPS target, estimated by quantifying the scene complexity.  

\begin{figure}
    \centering
    \includegraphics[width=0.99\linewidth]{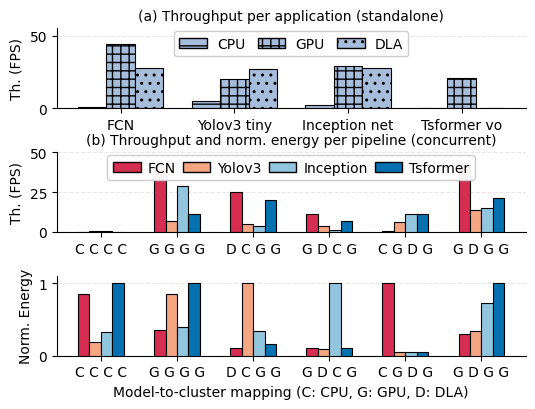}
    \caption{Throughput of perception module with different model-to-cluster mapping configurations.}
    \vspace{-12pt}
    \label{fig:back_4}
\end{figure}

Figure~\ref{fig:back_3} (b) and (c) present energy and throughput met rate with fixed (5/10/15) and adaptive FPS strategies on the Orin NX platform. Fixed-FPS strategies use a static model-to-cluster mapping across all regions. With fixed strategies, energy consumption increases monotonically with FPS. An adaptive strategy uses different FPS targets per region based on scene complexity, resulting in lower energy consumption while meeting throughput targets. The adaptive strategy's region-specific run-time model-to-cluster selection, with configurations: (e.g., Regions 1 and 3: 10 FPS mapped to (G: Yolov3 tiny, G: TSformerVO, D: FCN, D: inceptionet v3), Regions 2 and 3: 5 FPS mapped to (C: Yolov3 tiny, G: TSformerVO, D: FCN, G: inceptionet v3) and higher-load regions such as Region 4 requiring more heterogeneous mappings such as (G: Yolov3 tiny, G: TSformerVO, D: FCN, G: inceptionet v3)). It should be noted that the lower energy consumption of fixed (5/10 FPS) strategies in Regions 1, 3 and 4 is due to resource under-provisioning, without meeting appropriate throughput targets. In Region 2, fixed-15 FPS over-provisions by 3× relative to scene requirements, while fixed-5 FPS under-provisions in Regions 1, 3, and 4, where higher FPS is demanded, missing 18-75\% of throughput targets. Fixed-10 FPS partially balances this trade-off, matching Regions 1 and 3 but still missing 25–35\% in Region 4 and over-provisioning in Region 2. Although fixed-15 FPS achieves near-100\% throughput across all regions, it incurs 1.5-3x energy overhead in simpler regions (1–3). In contrast, the adaptive strategy leverages awareness of scene complexity to jointly adjust FPS and run-time model-to-cluster mapping, operating at low energy in simple regions such as Region 2, at moderate energy in Regions 1 and 3, and scaling only when necessary in Region 4. This results in 23–69\% energy savings over fixed-15 FPS while achieving a 100\% throughput-met rate, demonstrating that static configurations cannot co-optimize energy and performance under varying scene complexity. The actual per-region perception module orchestration challenges for an adaptive perception pipeline on \acp{hmp} are presented in Figure \ref{fig:back_4}, highlighting the feasibility of model deployment across clusters for varying FPS targets.

\begin{figure}
    \centering
    \includegraphics[width=0.99\linewidth]{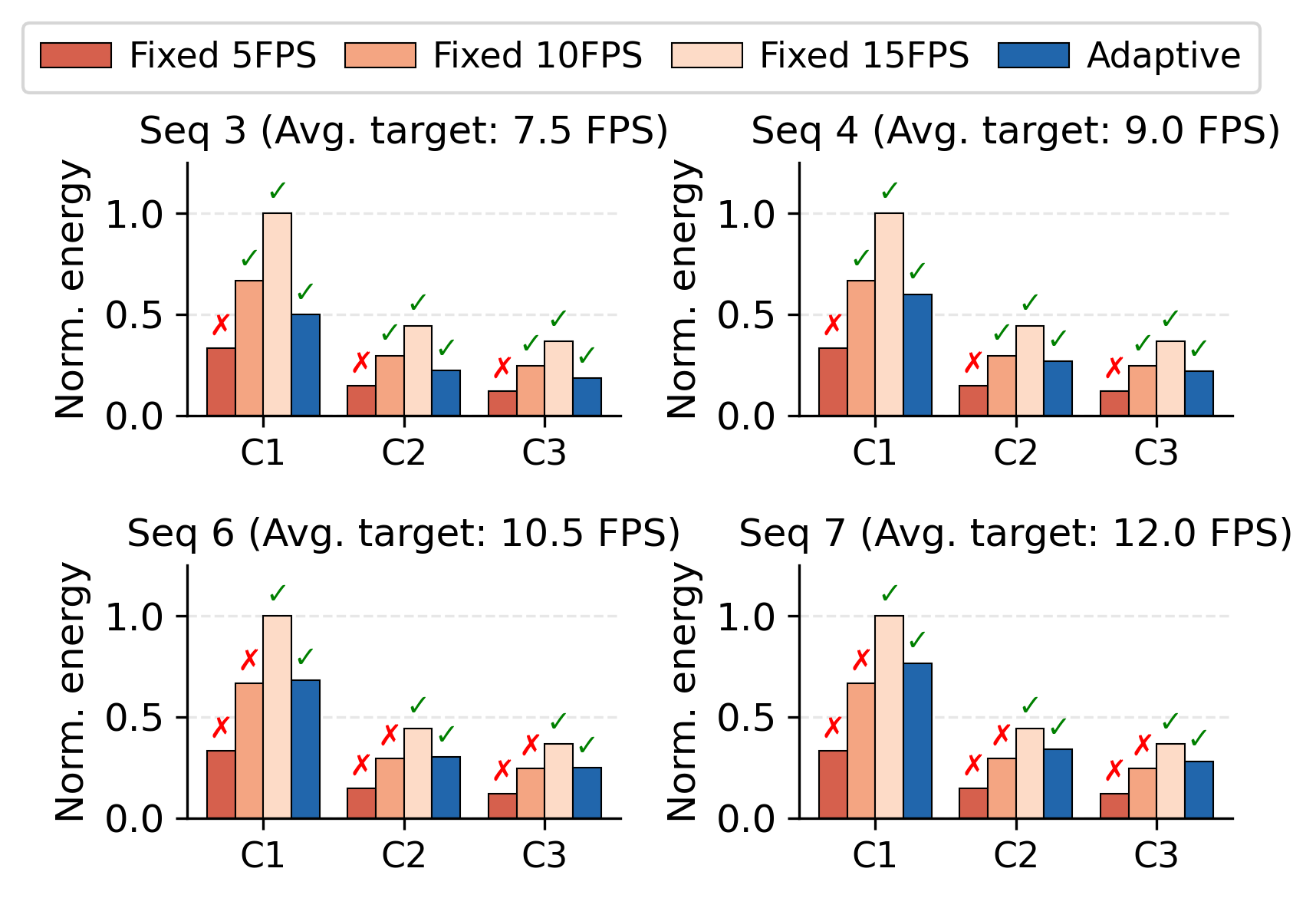}
    \caption{Energy comparison under fixed (5, 10, and 15 FPS) and variable throughput requirements. Mapping configs: C1: CPU, C2: CPU+GPU, C3: CPU+GPU+DLA
    }
     \vspace{-12pt}
    \label{fig:back_2}
\end{figure}

\subsubsection{Orchestrating adaptive perception on HMPs} \label{sec.orch_hmp}
We present challenges of orchestrating adaptive perception pipeline on \acp{hmp} from both the application and hardware perspectives under variable FPS requirements.

\begin{figure*}
    \centering
    \includegraphics[width=0.99\linewidth]{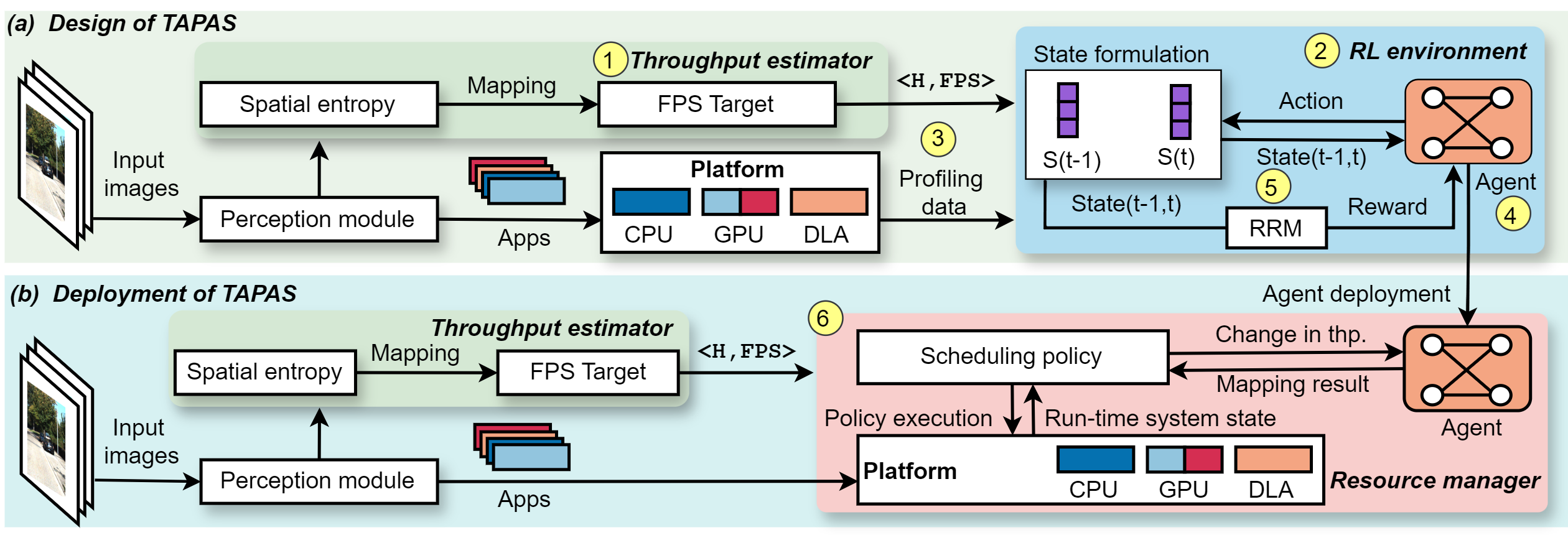}
    \caption{Training the GRU-based RL agent with offline profiled data generated using \textit{KITTI} trajectories.}
    \vspace{-12pt}
    \label{fig:TAPAS_sys_arch}
\end{figure*}

For the perception pipeline, we consider FCN, YoloV3Tiny, InceptionNet, and TransformerVO models, and Jetson Orin NX as the hardware platform. We execute the perception pipeline for 20 iterations on Jetson Orin NX and report the throughput and energy across runs. Figure~\ref{fig:back_4}(a) shows achievable throughput for each model on CPU, GPU, and DLA clusters when run in standalone mode. DNN-based models (FCN, YOLOv3-tiny, InceptionNet) achieve higher throughput on GPU while also achieving comparable throughput on DLA. In contrast, transformer-based TSFormerVO lacks operator support on DLA and is confined to the GPU. Lightweight YOLOv3-tiny shows balanced performance across GPU (20 FPS), DLA (27 FPS), and CPU (5 FPS), reflecting its flexibility in deployment. Figure~\ref{fig:back_4}(b) illustrates the impact of model-to-cluster mapping on throughput and energy, when multiple models are run concurrently. In this example, we map perception models to the CPU (C), GPU (G), and various combinations of CPU, GPU, and DLA (D). Mapping all workloads on CPU leads to severe throughput degradation ($<$1 FPS), and significantly higher energy consumption. GPU-only execution improves aggregate throughput, while yielding relatively lower normalized energy consumption for specific models. It should be noted that GPU-only execution is subject to memory constraints; running four models concurrently on a GPU can cause one or more processes to be killed due to insufficient memory and require rescheduling. DNN-based models benefit from being mapped onto DLA, achieving moderate-to-high throughput at lower energy consumption by exploiting accelerator hardware. With throughput-energy diversity across clusters and models, adapting to varying throughput targets while minimizing energy consumption poses a complex optimization challenge. This necessitates a dynamic model-to-cluster mapping that considers model-cluster affinity, operator support variability, and throughput-energy trade-offs.

\noindent \textbf{Platform- and scene-specific impact on adaptive perception.} While adaptive perception provides throughput and energy gains, the extent of such gains largely depends on variation in scene complexity and platform-specific hardware. 
To illustrate this, we use KITTI sequences 3, 4, 6, and 7 with varying average FPS targets, and the Jetson Orin NX platform with CPU, GPU, and DLA clusters. We consider a lightweight perception pipeline (e.g., YOLOv11n, ResNet-50, FCN-ResNet50) for different configurations.

Figure~\ref{fig:back_2} shows normalized energy with fixed (5, 10, 15 FPS) and adaptive perception strategies across C1 (CPU-only), C2 (CPU+GPU), and C3 (CPU+GPU+DLA) mapping configurations. A green tick indicates the throughput target is being met, and a red cross indicates it is not met. In seq 3, the average throughput target is 7.5 FPS, which is met with fixed 10 and 15 FPS, and adaptive perception strategies. Among these, the adaptive perception strategy has the lowest energy consumption across C1, C2, and C3 configurations. However, the extent of energy savings varies across clusters. Seq 4, with an average throughput target of 9 FPS, follows a similar trend to that of seq 3. It should be noted that the energy savings of the adaptive perception strategy are relatively lower as the average throughput target increases from 7.5 to 9 FPS between seq 3 and seq 4. Effectively, the resource wastage of the fixed 10 FPS strategy has lowered between seq 3 and seq 4, reflecting in lower energy gains with adaptive perception in this specific case. In seq 6 and seq 7, the throughput targets are 10.5 and 12 FPS, and only the fixed 15 FPS and adaptive perception strategies meet them. In this case, adaptive perception has lower overall energy consumption, while the magnitude of energy savings is similar to that in the previous sequences.

In summary, adaptive perception on \acp{hmp} faces three core challenges: (i) developing a mechanism to capture scene complexity and estimate variable FPS targets, (ii) a lightweight run-time system to configure model-to-cluster mapping based on variable FPS targets, and (iii) balancing the trade-off between throughput and energy, considering application and hardware diversity. Based on these motivations, we design \textit{TAPAS} framework for adaptive perception on \acp{hmp}. \textit{TAPAS} incorporates a scene-aware mechanism to estimate variable FPS targets, a lightweight agent for dynamic mapping to meet variable FPS targets, and a structured reward design for stable and grounded decision-making. The next section details the \textit{TAPAS} framework.

\section{TAPAS Framework} \label{sec.fraemwork}

 \subsection{Framework Overview} We design \textit{TAPAS} as a generic framework for orchestrating a throughput-adaptive perception pipeline on mobile \acp{hmp}. Figure \ref{fig:TAPAS_sys_arch} shows an overview of the \textit{TAPAS} framework, split into design and deployment phases. 
 \newline \noindent \textbf{Design phase.} In the design phase, we train an \ac{rl} agent to determine model-to-cluster mapping for variable FPS requirements. Initially, we profile the perception pipeline on an \ac{hmp} over different training sequences to collect execution traces of per-cluster latency, system-wide energy consumption, and an overall application profile. The \textit{ \ding{202} throughput estimator} module determines spatial entropy of images from the training data, and maps the entropy values to variable FPS targets. We create an \textit{\ding{203} \ac{rl} environment} with \textit{\ding{204} profiling data}, including the execution traces and spatial entropy-FPS values. We train a Gated Recurrent Unit (GRU)-based scheduling agent within a \ac{ppo} \ac{rl} framework, where \ac{ppo} ensures stable policy updates under dynamic throughput variations. We enhance the efficiency of the RL agent through a \textit{\ding{206} Reward Reasoning Model (RRM)}, which provides structured, context-aware reward signals based on throughput-met rate, energy efficiency, and entropy-driven constraints. The recurrent \textit{\ding{205} GRU} architecture learns to map temporal state sequences to optimal cluster configurations that minimize energy consumption while satisfying throughput targets. The hidden state encodes patterns of entropy and workload history across sliding windows. 
 \newline \textbf{Deployment phase.} We monitor changes in throughput requirements at run-time while \textit{\ding{207} deploying}, and enforce model-to-cluster mapping decisions determined by the RL agent. At each frame $t$, we compute spatial entropy from object detection outputs (within the perception pipeline), assign a variable FPS target based on scene-complexity, and construct a state vector via n-frame temporal stacking. The \ac{gru} agent processes this state using its gating mechanisms, updates its hidden state, and makes a run-time model-to-cluster mapping decision.

 \vspace{-9pt}

\subsection{Problem formulation}
We formulate throughput-adaptive perception for \ac{as} on mobile \acp{hmp} as a resource allocation problem. The perception pipeline consists of a set of heterogeneous workloads $\mathcal{W}={w_1, w_2, \ldots, w_N}$, representing deep neural networks and transformer-based models for tasks such as detection, segmentation, and tracking. These workloads are executed on an \ac{hmp} platform with multiple compute clusters $\mathcal{C}={C_1, C_2, \ldots, C_M}$, each exhibiting distinct performance–energy trade-offs. At run-time, the system is subject to variable throughput requirements expressed as target frame rates $T_v \in {T_{v1}, T_{v2}, \ldots, T_{vn}}$, which fluctuate with environmental dynamics. The goal is to learn an optimal scheduling policy $\pi(s_t)=a_t$ that maps the system state at time $t$ to an action $a_t$, where the action assigns each workload $w_i$ to a cluster $\mathcal{C}$ such that throughput target is met within lowest energy consumption. 

The objective of the TAPAS policy is to jointly minimize the throughput deficit and the system-wide energy consumption. This is formalized as the following optimization problem:
\begin{equation}
\min_{\{h_{i,t}\}_{i=1}^{N}} \; O_t =
\left( T_v(t) - \sum_{i=1}^{N} R_i(h_{i,t}) \right)
\;+\;
\left( \sum_{c \in \mathcal{C}} E_{sys}(u_{c,t}) \right),
\label{eq:objective}
\end{equation}
where $\sum_{i=1}^{N} R_i(h_{i,t})$ denotes the aggregate achieved throughput, and $\sum_{c \in \mathcal{C}} E_{sys}(u_{c,t})$ captures the total energy consumed by all clusters under utilization state $u_{c,t}$. In Equation \ref{eq:objective}, the first term penalizes deviations between the target and achieved FPS, while the second term penalizes excessive energy usage. By minimizing this combined objective, the scheduler delivers the target throughput with lower energy consumption. 

\subsection{Design of TAPAS} 

\noindent \subsubsection{\textit{\textbf{Throughput estimator}}} \label{sec.spatial_entropy} The target FPS (throughput) is determined for a given frame based on the scene complexity. We consider the perception pipeline to implicitly include an object detection model, and use its output to determine spatial entropy, which quantifies scene complexity based on the number of detected objects. Since perception must complete within a fixed time window, different frame complexities translate into variable throughput requirements. TAPAS exploits this relationship by mapping spatial entropy ranges to FPS targets at run-time. For each frame $I_t$, the object detection model $\mathcal{O}$ obtains a semantic class map $C_t$, which encodes the spatial distribution of objects in the scene. A normalized histogram $\mathbf{p}_t$ is computed over $C_t$, and spatial entropy is derived via Shannon's formulation:
\begin{equation}
\label{eq:spatial_entropy}
    H_t = -\sum_{c} p_t(c) \log p_t(c)
\end{equation}

The number of levels $N_h$ and $N_t$ directly control the number of levels for spatial entropy and FPS target, while the class-gap parameters $CG_t$ and $CG_h$ govern the granularity of both spatial entropy and FPS target, respectively. Both the class gap and the number of levels are fixed design-time parameters, with $(N_h, N_t, CG_t, CG_h)  \in \{1, 2, \ldots, N\}$ configured before deployment based on the target platform's FPS operating range and expected scene complexity distribution. These parameters are necessary because a single fixed FPS target cannot capture the continuous variation in scene complexity, yet exhaustive per-frame FPS optimization is computationally infeasible at runtime; the class-gap and level count together provide a lightweight discretizations that balances adaptation granularity against scheduling overhead. Given a baseline entropy $H_{base}$, the entropy thresholds for $N_h$ levels are parameterized as:
\begin{equation}
    \label{eq:entropy_threshold}
    H_k = H_{base} + (k - \lceil N_h/2 \rceil) \cdot CG_h, 
    \quad k = 1, 2, \dots, N_h
\end{equation}
where $CG_h$ defines the bandwidth between consecutive entropy levels. Once the $N_h$ entropy bands are established, the throughput estimator maps each frame's entropy $H_t$ to a variable FPS target $T_v$ across $N_t$ throughput levels, spaced by $CG_t$ relative to baseline $T_{base}$:
\begin{equation}
\label{eq:fps_target}
    T_{vi} = T_{base} + (j - \lceil N_t/2 \rceil) \cdot CG_t, 
    \quad j = 1, 2, \dots, N_t
\end{equation}
such that $H_t \in [H_{k}, H_{k+1})$ maps to the $j$-th throughput level $T_{vi}$. Increasing $N_h$ and $N_t$ enables finer-grained runtime adaptation, while larger $CG_h$ and $CG_t$ widen the entropy bands and the FPS target steps, respectively, providing independent control over sensitivity to scene complexity and responsiveness to throughput.

\subsubsection{\textit{\textbf{Design of RL environment}}} \label{sec.rl}
Training the TAPAS RL agent requires a well-defined environment comprising three interdependent components: a state space, an action space, and a reward model. This section details the \ac{rl} environment design, \ac{gru}-based agent architecture, and \ac{rrm}, and describes how these components are jointly trained to enable adaptive cluster mapping. We process \textit{KITTI} trajectory image sequences over perception workloads $\mathcal{W}$ on a representative \ac{hmp} with clusters $C_n$ to calculate spatial entropy and corresponding variable throughput requirements. Overview of the training process is shown in Figure \ref{fig:TAPAS_sys_arch}. As mentioned in Section \ref{sec.spatial_entropy}, spatial entropy quantifies dynamicity within image sequences \cite{Shannon_spatial_entropy} using Shannon's formula (Eq. \ref{eq:spatial_entropy}). We measure throughput metrics for each workload configuration on all the \ac{hmp} clusters $\mathcal{C}$.

\begin{figure}
    \centering
    \includegraphics[width=0.99\linewidth]{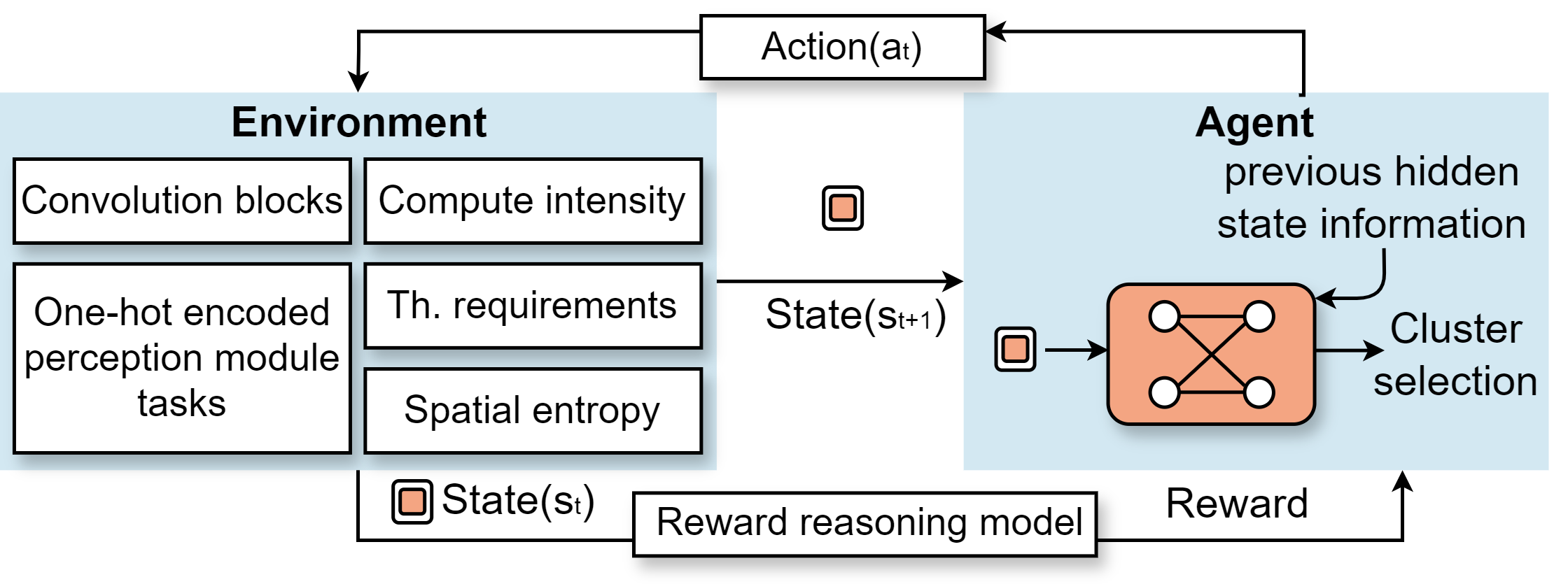}
    \caption{RL environment with GRU agent and reward reasoning model.}
    \vspace{-15pt}
    \label{fig:rl_env}
\end{figure}

\noindent \textit{In-context state space formulation.}
Overview of the RL environment with GRU agent and \ac{rrm} is shown in Figure \ref{fig:rl_env}. At decision step $t$, the RL environment processes perception workloads $\mathcal{W} = \{w_1, w_2, ..., w_N\}$ where each workload $w_i$ is characterized by:

\begin{equation}
\phi_i =
\begin{aligned}
[\text{GFLOPs}_i, \text{ConvBlocks}_i, \text{TransformerEnc}_i,\\
\text{MultiHeadAttn}_i, \text{TaskType}_i, T_{vi}, H(S_t)]
\end{aligned}
\in \mathbb{R}^d
\end{equation}

This captures computational intensity, architectural complexity (convolution/transformer blocks, multi-head attention), one-hot encoded perception tasks, variable throughput requirements $T_{vi}$, and spatial entropy $H(S_t)$.The global state aggregates all workloads as 
$s_t = [s^t_1, s^t_2, \dots, s^t_N] \in \mathbb{R}^{dN}$, where each $s^t_i$ captures the application characteristics, current variable FPS target $T_t$ and $H(S_t)$ for each perception workload $i$ in the pipeline. To enable in-context state formulation, the current and previous state vectors are 
concatenated as $\tilde{s}_t = [s_{t-n} \| s_t] \in \mathbb{R}^{2dN}$, providing the GRU agent with temporal context to capture workload transition patterns and entropy evolution across consecutive frames.

\noindent \textit{Action Space:} The RL agent's action space includes the selection of a cluster for perception workloads:

\begin{equation}
    a_t = \{c^t_1, c^t_2, \dots, c^t_N\}, \quad c^t_i \in \mathcal\{\text{Compute cluster}\}
\end{equation}

\subsubsection{\textit{\textbf{Agent architecture}}} \label{sec.gru}
The GRU agent maintains an in-context state representation by concatenating $s_{t-1}$ and $s_t$, capturing temporal dependencies across consecutive scheduling decisions. At each timestep $t$, the state vector $[H_t, T_v, \phi_t]$ encoding spatial entropy, target FPS, and workload characteristics is fed into the GRU hidden state, enabling the agent to track entropy evolution, throughput stability, and workload transition patterns across frames. This recurrent formulation allows the agent to anticipate shifts in scene complexity (e.g., from sparse highway to dense intersection) and adapt cluster assignments accordingly, directly addressing the fundamental limitation of static scheduling methods that lack historical context.

The \ac{gru} \cite{GRU} employs reset and update gates that selectively retain relevant historical information while filtering out obsolete scheduling decisions, enabling effective learning from past workload allocation experiences. The reset gate $r_t = \sigma(W_r \cdot [h_{t-1}, s_t])$ determines which previous scheduling patterns to forget, while the update gate $z_t = \sigma(W_z \cdot [h_{t-1}, s_t])$ controls the integration of new state information with historical context. This temporal memory mechanism enables the agent to recognize recurring throughput patterns, adapt to cyclical variations in throughput, and make informed scheduling decisions based on both the current system state and the hidden state of past allocation outcomes. 

\subsubsection{\textit{\textbf{Reward Reasoning Model}}} \label{sec.rrm}

\begin{figure}
    \centering
    \includegraphics[width=0.99\linewidth]{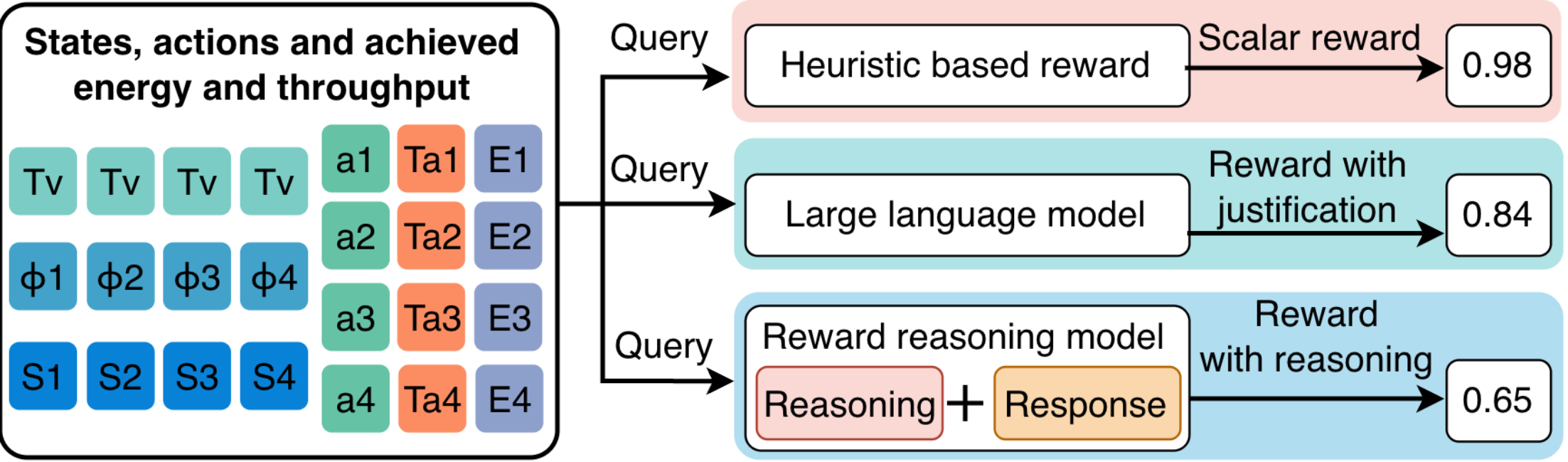}
    \caption{RL-based reward shaping approaches: heuristic-based reward, LLM-based reward, and  Reward reasoning model.}
    \vspace{-18pt}
    \label{fig:rrm}
\end{figure}

 This subsection details the \ac{rrm} employed within the TAPAS framework to compute structured, multi-objective rewards for \ac{rl} agent training. 
\ac{rl} agents suffer from reward shaping problems \cite{reward_shaping}. Figure \ref{fig:rrm} compares three reward shaping approaches for \ac{rl}-based scheduling using system features (variables throughput requirement $P_v$, workload specific states $\phi_i$, S, actions $a_i$, metrics $P_a$, $E_a$). Heuristic-based rewards \cite{tango,omniboost} achieve the highest confidence (0.98) without explanation; they rely on simplistic linear combinations of throughput and energy metrics, leading to suboptimal \ac{rl} policy behavior. \ac{LLM}-based rewards \cite{LLM_based_reward} justify moderate scores (0.84), but lack domain-specific reasoning for \acp{hmp} and variable FPS targets. \ac{rrm} \cite{reasoning_reward_model} represents a novel approach that leverages the Qwen2 \cite{qwen} Transformer-decoder architecture, offering detailed explanations with verifiable rewards (0.65). Unlike traditional reward functions, \ac{rrm} formulates reward modeling as a text completion problem, auto-regressively generating comprehensive output consisting of structured thinking processes followed by a final response. This enables chain-of-thought reasoning and provides structured, interpretable feedback that dynamically evaluates scheduling decisions based on contextual factors (e.g., spatial entropy, throughput requirements, and hardware constraints) rather than on fixed weight assignments. This approach helps the GRU agent understand \textit{why} certain allocations are beneficial beyond mere reward, enabling stabler learning and faster convergence. Further, it provides better generalization to variable FPS targets while maintaining energy efficiency across \acp{hmp} compared to traditional heuristic approaches. A potential risk of \ac{rrm} is hallucination, where generated reasoning may diverge from system reality \cite{hallucination_llm_survey}. To address this, we use grounded reward with \ac{rrm} to ensure that only factually consistent reasoning is propagated to the GRU agent.

Standard LLM-based reward models compute a scalar score from textual descriptions of states and actions. These approaches rely mainly on human preferences or supervised labels, without grounding predictions in real system behavior. In contrast, the \ac{rrm} conditions reward prediction on both structured reasoning traces and physical measurements (throughput, energy, entropy), and is trained with a grounding objective to align outputs with measured metrics.
A conventional LLM reward model computes a reward as:
\begin{equation}
R_{\text{LLM}}(s_t,a_t) \;=\; g_{\phi}\big(\mathrm{TextEncode}(s_t,a_t)\big),
\label{eq:llm_reward}
\end{equation}
emphasize linguistic priors but lack explicit grounding in real-world outcomes.

We define a \textit{grounded reward} as a physically constrained and measurement-consistent supervision signal for reinforcement learning. The \textit{grounded reward} is explicitly derived from real-world system outcomes rather than from heuristic approximations or unconstrained language-model outputs. Formally, at decision step $t$, given the system state $s_t$, the selected scheduling action $a_t$, and the vector of offline measured platform outcomes $\mathcal{M}_t$ (including achieved throughput, latency, and energy), the grounded reward $\hat{R}_t$ is generated as
\begin{equation}
    \hat{R}_t = f_{\theta}(s_t, a_t, \mathcal{M}_t),
\end{equation}
where $f_{\theta}(\cdot)$ denotes the \ac{rrm} parameterized by $\theta$. Unlike conventional LLM-based reward models that score plausibility without grounding, RRM conditions rewards on measured outcomes $\mathcal{M}_t$, penalizes predictions that contradict real observations via a grounding loss, and applies adaptive test-time reasoning to verify candidate reward estimates where uncertainty is highest, replacing free-form scoring with a measurement-anchored reward reshaping.

\begin{algorithm}
\caption{Agent training}
\label{alg:gru_training}
\footnotesize
\begin{algorithmic}[1]
\REQUIRE Offline image sequences $\mathcal{I}$, perception workloads $\mathcal{W}$, clusters $\mathcal{C}$, 
GRU policy $\pi_{\theta}$, RRM reward model $\mathcal{R}$, throughput estimator $\Psi(\cdot)$
\ENSURE Trained GRU scheduling policy $\pi_{\theta^{*}}$

\STATE Initialize GRU parameters $\theta$, replay buffer $\mathcal{B}$, baseline FPS $T_{base}$

\FOR{each trajectory $\tau$ in dataset}
    \FOR{each frame $I_t$ in trajectory}
        \STATE $(T_v) \leftarrow \Psi(I_t, \mathcal{O}, T_{base},N_h, N_t, CG_t, CG_h)$ 
        \COMMENT{Call throughput estimator}

        
        \STATE Extract workload features 
        \[
        \begin{aligned}
        \phi_i \leftarrow [&\text{GFLOPs}_i, \text{ConvBlocks}_i,\text{TransformerEnc}_i, \text{MHA}_i,  \\
                  &\text{TaskType}_i, T_v, H(S_t)]
        \end{aligned}
        \]

        \STATE Construct global state $s_t = [\phi_1, \phi_2, ..., \phi_N]$

        \STATE Sample action $a_t \sim \pi_{\theta}(a|s_t)$ 
        \COMMENT{Assign each workload to a compute cluster}

        \STATE Execute scheduling action $a_t$ on clusters $\mathcal{C}$, measure throughput $P_a$ and energy $E_a$

        \STATE Compute reward $R_t \leftarrow \mathcal{R}(s_t, a_t, P_a, E_a, T_t)$
        \COMMENT{Reward reasoning model}

        \STATE Store transition $(s_t, a_t, R_t, s_{t+1})$ into $\mathcal{B}$
    \ENDFOR

    \STATE Update GRU parameters $\theta \leftarrow \theta - \alpha \nabla_{\theta}L_{\text{PPO}}(\theta, \mathcal{B})$
    \COMMENT{Policy optimization with PPO}
\ENDFOR

\RETURN Trained GRU policy $\pi_{\theta^{*}}$
\end{algorithmic}
\end{algorithm}

\subsubsection{\textit{\textbf{RL agent training}}}
GRU-based agent learns to dynamically map perception models to heterogeneous compute clusters within a closed training loop, which comprises of throughput estimator, offline profiled data, RL environment, and \ac{rrm}. Algorithm~\ref{alg:tapas_training} summarizes the end-to-end offline training procedure for the GRU-based RL agent.

\noindent\textit{Agent training algorithm}: The GRU scheduling agent is trained offline using trajectory data from KITTI and throughput labels derived from the throughput estimator. The TAPAS agent training algorithm takes as inputs: offline image sequences $\mathcal{I}$, perception workloads $\mathcal{W}$, clusters $\mathcal{C}$, GRU policy $\pi_{\theta}$, RRM reward model $\mathcal{R}$, throughput estimator $\Psi(\cdot)$. The training loop begins by initializing GRU parameters, and the baseline FPS setting $T_{base}$ (Line 1). For each trajectory $\tau$ in the dataset (Line 2), and each frame $I_t$ within that trajectory (Line 3), throughput estimator $\Psi(\cdot)$ is invoked to compute variable FPS target $T_v$ with arguments $(T_{base},N_h, N_t, CG_t, CG_h)$\textit{ (Line 4)}. These serve as ground-truth supervisory signals for variable throughput scheduling. The agent then extracts workload descriptors $\phi_i$ that capture compute intensity, architectural composition, task type, throughput target $T_v$, and the spatial entropy $H(S_t)$ (Line 5). These descriptors are aggregated into a global state vector $s_t$ (Line 6) representing the complete scheduling context at time $t$. Using this state representation, the GRU policy samples an action $a_t$ (Line 7), which corresponds to assigning optimal model-to-cluster mapping. The action is executed on the \ac{hmp} (Line 8), allowing measurement of achieved throughput $P_a$ and energy consumption $E_a$. The reward reasoning model evaluates the scheduling action (Line 9) and generates a structured, context-aware reward $R_t$ that accounts for throughput demands, energy efficiency, entropy-driven constraints, and architecture-specific performance variations. Each transition tuple is stored in a replay buffer for subsequent optimization (Line 10). After completing a trajectory, the GRU parameters are updated using PPO's clipped surrogate loss (Line 11), enabling stable policy improvement under variable-throughput conditions. This process iterates over all trajectories (Line 2), resulting in a trained GRU scheduling policy $\pi_{\theta^{*}}$ (Line 12).

\subsection{Agent deployment and integration in scheduling policy}

As presented in figure \ref{fig:TAPAS_sys_arch}, The resource manager consists of two tightly coupled components: (i) the scheduling policy and (ii) the RL-based GRU agent. The scheduling policy continuously monitors run-time system status, including cluster availability, workload characteristics, and current throughput demand, and determines when re-allocation is required. The GRU agent is invoked by the scheduling policy to compute optimal model-to-cluster mappings under the observed system state. As shown in Figure~\ref{fig:TAPAS_sys_arch}, this section details the deployment stage of TAPAS, wherein the trained GRU agent is integrated into the run-time scheduling policy as a core component of the resource manager, which enforces adaptive model-to-cluster mapping decisions directly onto the targeted platform. The scheduling policy triggers agent queries whenever variations in spatial entropy indicate shifts in FPS requirements, thereby enabling run-time workload-cluster mapping decisions. The lightweight GRU architecture minimizes computational overhead while providing optimal resource allocation for \ac{hmp} based on current perception-workload characteristics and variable throughput demands.

\begin{algorithm}
\caption{TAPAS Deployment Policy}
\label{alg:tapas_training}
\footnotesize
\begin{algorithmic}[1]
\REQUIRE Perception workloads $\mathcal{W}$, clusters $\mathcal{C}$, 
         throughput estimator $\Psi(\cdot)$, trained GRU policy $\pi_{GRU}$, 
         entropy threshold $\tau$
\ENSURE  Adaptive cluster mapping $M: \mathcal{W} \rightarrow \mathcal{C}$
\STATE Initialize $T_v^{curr} \leftarrow T_{base}$ FPS
\WHILE{system active}
    \STATE Observe run-time system status $z_t \leftarrow \{Ca_{t}\}$
    \STATE Compute spatial entropy $H(S_t)$ from current frame $I_t$
    \STATE $(T_v) \leftarrow \Psi(I_t, \mathcal{O},I_t, T_{base},N_h, N_t, CG_t, CG_h)$
    \COMMENT{Throughput estimator}
    \STATE Determine new throughput requirement $T_v^{new} \leftarrow f(H(S_t))$
    \IF{$|T_v^{new} - T_v^{curr}| > \Delta$}
        \STATE Extract workload features 
        \[
        \begin{aligned}
        \phi_i \leftarrow [&\text{GFLOPs}_i, \text{ConvBlocks}_i,\text{TransformerEnc}_i, \text{MHA}_i,  \\
                  &\text{TaskType}_i, T_t, H(S_t)]
        \end{aligned}
        \]
        \STATE Construct state $s_t \leftarrow [\phi_1, \phi_2, \ldots, \phi_N, T_v^{new}, H(S_t)]$
        \STATE Query GRU agent: $a_t \leftarrow \pi_{GRU}(s_t)$
        \STATE Execute cluster mapping: $\Pi_{SP} \leftarrow a_t$
        \STATE Update $T_v^{curr} \leftarrow T_v^{new}$
    \ENDIF
    \STATE Execute perception workloads $\mathcal{W}$ on assigned clusters through $\Pi_{SP}$
\ENDWHILE
\end{algorithmic}
\end{algorithm}

Algorithm~\ref{alg:tapas_training} presents the TAPAS run-time scheduling policy deployed on the target \ac{hmp}.The system initializes $T_v^{curr}$ with a baseline throughput target of $T_{base}$ FPS (Line 1) and enters continuous monitoring (Line 2). At each time step, the scheduling policy monitors run-time system status (cluster availability) (Line 3). At each frame, spatial entropy $H(S_t)$ is computed (Line 3) and passed to the throughput estimator $\Psi(\cdot)$ take arguments $(T_{base},N_h, N_t, CG_t, CG_h)$, producing variable throughput (Line 5). The new FPS requirement $T_v^{new}$ is derived from $H(S_t)$ (Line 6), and a remapping decision is triggered only when a change in throughput target is detected $\Delta$ (Line 7), minimizing scheduling overhead during steady-state operation. Upon a shift, the policy extracts per-workload features with  task type alongside current entropy and throughput $\phi_i$ (Line 8) and assembles the global state vector $s_t$ (Line 9). The trained GRU agent is then queried to produce the optimal cluster assignment $a_t$ (Line 10), which is executed as the updated scheduling policy mapping $\Pi_{SP}$ (Lines 11-12). Perception workloads are subsequently dispatched to their assigned clusters through scheduling policy $\Pi_{SP}$ (Line 13), and the loop continues monitoring for subsequent entropy-driven throughput changes.

\section{Experimental Evaluation}\label{sec.results}

\subsection{Experimental setup} \label{sec.exp_setup}
\noindent \textit{Platform.} \label{sec.platform}
We evaluate TAPAS on a representative mobile \ac{hmp} platform of Nvidia Jetson Orin NX ~\cite{orin_nx}, featuring an Ampere GPU with 1024 CUDA cores (765 MHz max), a hexa-core ARM Cortex-A78 CPU (1.9 GHz max), and 8 GB shared memory with asymmetric CPU clusters (4+2 ARM cores). The TAPAS framework is cross-platform, portable middleware that generalizes across scenes and perception pipelines, although the achievable gains depend on the target platform.

\noindent \textit{Workloads.} \label{sec.workload}
For the perception module, we use SOTA models spanning object detection (YOLOv3-tiny, YOLOv11n, Faster R-CNN), semantic segmentation (FCN, DeepLab V3), visual odometry (TSFormerVO, DeepVO), and obstacle avoidance (ViT, Swin Transformer, InceptionNet V3), covering diverse compute patterns such as convolution, attention, and encoder–decoder architectures. For evaluation, we set throughput estimator parameters as $N_h = N_t = 3$, $H_{base} = 1.5$, $CG_h = 1.0$, $T_{base} = 10$ FPS, and $CG_t = 5$ FPS.

\noindent \textit{Workloads Mixes.}
For experimental evaluation, we use five application mixes with varying compute intensity, based on eleven perception models: M0 (Faster R-CNN), M1 (Deeplab v3), M2 (Swin transformer), M3 (YOLOv11-n), M4 (FCN), M5 (TSformer VO), M6 (ViT), M7 (YOLOv3 tiny), M8 (InceptionNet), M9 (ResNet-152), and M10 (DeepVO). Application mixes contain moderate to high complexity perception modules -- Mix 1: heavier detection and segmentation (M0, M1, M2), Mix 2: lightweight detection and transformer-based obstacle detection (M3, M4, M5), Mix 3: detection, segmentation, and visual odometry (M0, M4, M5), Mix 4: lightweight detection, and obstacle avoidance and heavier segmentation, and visual odometry (M7, M4, M8, M5), and  Mix 5: heavier detection, obstacle avoidance, segmentation, and visual odometry. (M0, M4, M5, M8)

\noindent \textit{Dataset.}
For training, we use \textit{KITTI} dataset sequences (0, 1, 9, 8,10)  \cite{kitti_dataset}, which comprises diverse autonomous driving scenarios with varying traffic densities and lighting conditions. 
For evaluation and testing, we use KITTI dataset sequences (2, 3, 4, 5, 6, 7) and all the sequences of nuScenes dataset~\cite{nuscenes}. Our evaluation on test sequences from both KITTI and nuScenes datasets that are unseen by the RL agent validates the generalizability of the TAPAS framework for providing adaptive perception across diverse real-world scenarios.

\noindent \textbf{Comparison w.r.t. SOTA.}
For evaluation, we consider three SOTA perception and multi-DNN execution strategies with varying degrees of hardware heterogeneity exploration: \textit{EE} \cite{ee_tro} (CPU), \textit{Omniboost} \cite{omniboost} (CPU+GPU), and \textit{Band} \cite{band_npu} (CPU+GPU+NPU). For EE \cite{ee_tro}, we implement power, energy, and performance models by profiling applications and incorporating them into heuristic decision policies. For \textit{OmniBoost}, we use Gymnasium for implementing the Monte Carlo Tree Search algorithm and PyTorch for training ResNet-based agents on our profiled dataset. \textit{Band} partitions the perception module's models into subgraphs at design time, and maps supported DNN operations onto \ac{hmp}. We implemented the \textit{Band}’s model analyzer for subgraph generation. We measure the throughput of each vision task in the perception module, including run-time variable throughput requirements and energy across diverse workload mixes.

\subsection{Ablation studies of RL agent and RRM} \label{sec.ablation_rrm_gru}
\noindent \textit{(a). RL training.}
Our GRU-based agent utilizes \ac{ppo} \cite{PPO} as the \ac{rl} training algorithm, complemented by a novel \ac{rrm} \cite{reasoning_reward_model} for reward shaping. We train our agent on two experimental setups: the first containing 3 models and the second with 4 models on \textit{KITTI} dataset sequences (0, 1, 9, 8,10) with remaining sequences (2, 3, 4, 5, 6, 7) reserved for testing, demonstrating robust scheduling behavior and adaptability in unseen test environments. We train on a continuous stream of input frames from the KITTI dataset. In our evaluation, we use three FPS levels (5, 10, and 15 FPS) based on empirically observed spatial entropy levels across the KITTI dataset (Figure \ref{fig:rpp}). However, our framework is designed to support multiple FPS levels (Section \ref{sec.fraemwork}), offering fine-grained control over adaptive perception. We emulate variable throughput by skipping an appropriate number of frames (as proposed in~\cite{slim_slam}) to achieve specific FPS targets (5, 10, 15).

\begin{figure}
    \centering
    \includegraphics[width=0.99\linewidth]{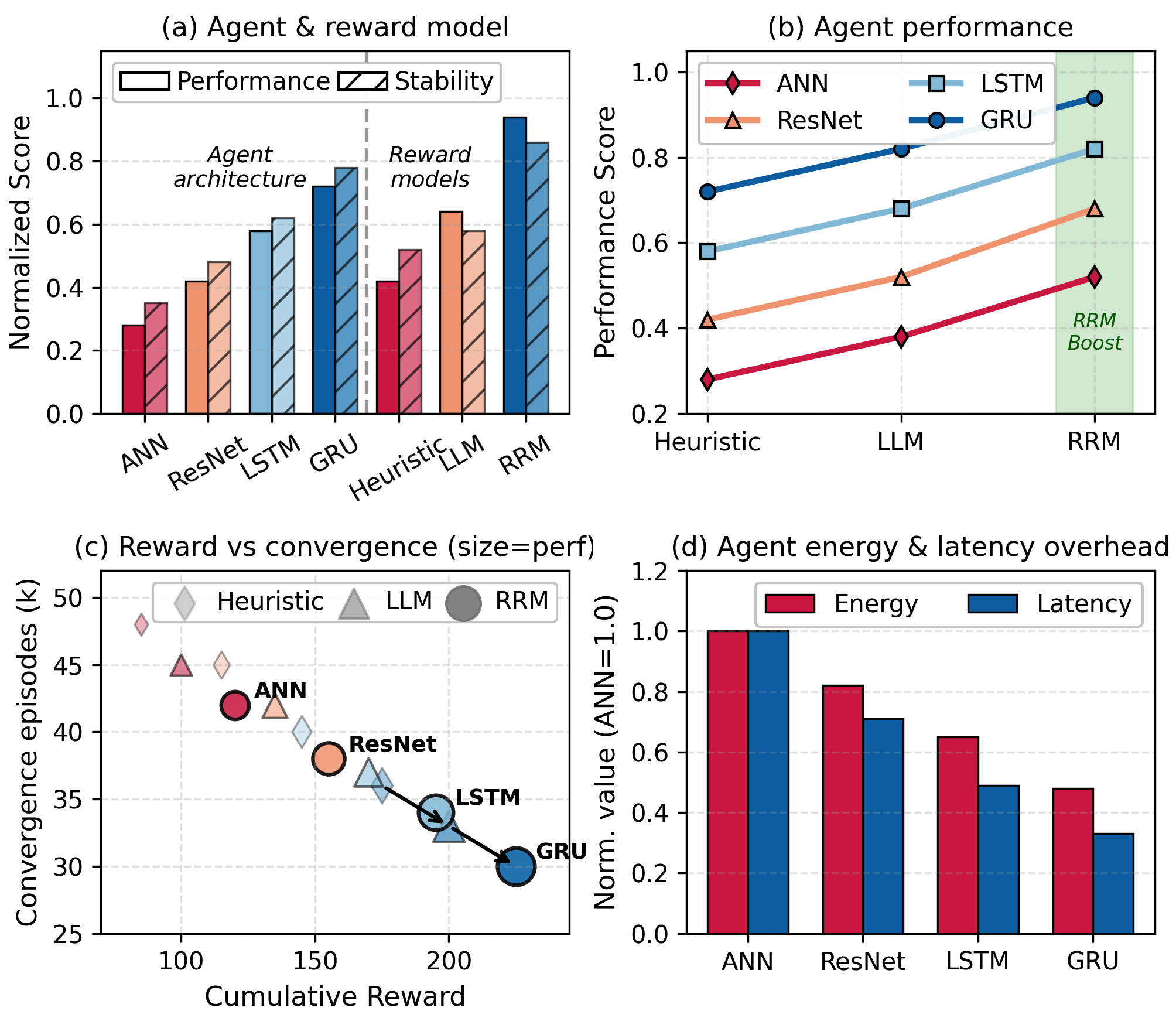}
    \caption{Ablation study of RRM and GRU-based agent in terms of performance, stability, convergence, and computational cost.}
    \label{fig:gru_performance_analysis}
    \vspace{-12pt}
\end{figure}

\noindent \textit{(b). Agent and reward model design space evaluation.}
Figure ~\ref{fig:gru_performance_analysis}(a) presents an ablation study across agent architectures (ANN \cite{tango}, ResNet \cite{omniboost}, LSTM \cite{RELMAS_lstm_dac}, GRU) and reward models (Heuristic, LLM, \ac{rrm}), showing the importance of temporal modeling and grounded rewards for \textit{TAPAS}. Among these agents, GRU consistently outperforms others, achieving the highest performance (0.72) and stability (0.78). ANN-based agents lack memory, while ResNet agent only captures spatial features. An LSTM agent improves temporal learning but incurs higher overhead. GRU achieves the best trade-off through a lightweight two-gate design that efficiently encodes workload history and adapts rapidly to variable FPS. Across reward models, heuristic rewards under-perform due to their handcrafted nature, which yields sparse learning signals. LLM-based rewards improve semantic reasoning, but suffer from stochasticity and inconsistency across similar state-action pairs. \ac{rrm} achieves the highest performance and stability (0.94/0.86) by producing grounded and deterministic rewards via test-time reasoning over spatial entropy, throughput, energy, and adaptivity. This structured reasoning reduces exploration variance by 62\% and enables reliable convergence, making GRU+RRM combination particularly effective for adaptive perception scheduling across \acp{hmp}.

\noindent \textit{(c). Model ablation across agent architectures.}
Figure \ref{fig:gru_performance_analysis}(b) shows consistent performance improvement from Heuristic$\rightarrow$LLM$\rightarrow$RRM across all agents: (i) ANN (+86\%), (ii) Resnet (+62\%), (iii) LSTM (+41\%) and (iv) GRU (+31\%). Heuristic$\rightarrow$LLM transition provides moderate gains by incorporating semantic context, though improvements plateau due to stochasticity. The LLM$\rightarrow$RRM transition yields greater improvement through adaptive test-time reasoning, eliminating variance through deterministic evaluation. The diminishing improvement percentage (86\%→31\%) indicates that simple agents (such as ANN) benefit more from improved rewards (\ac{rrm}), while advanced agents (such as LSTM and GRU) already capture temporal dynamics effectively. Notably, ANN+RRM (0.52) approaches ResNet+Heuristic (0.42), indicating that reward quality partially compensates for architecture. Whereas, GRU+Heuristic (0.72) exceeds ANN+RRM (0.52), confirming that optimal performance requires both strong architectures and high-quality reward modeling.

\noindent \textit{(d). Convergence and cumulative reward analysis for agent architecture and reward model.} 
Figure \ref{fig:gru_performance_analysis}(c) illustrates progressive improvement across reward models; marker size encodes performance and arrows trace learning trajectories. GRU improves from 175 reward/36k episodes (Heuristic) $\rightarrow$ 225 reward/30k episodes (RRM), achieving 29\% higher reward with 17\% faster convergence. Heuristic reward stuck in a suboptimal region (36-48k episodes, 85-175 reward) due to sparse binary rewards, leading to inefficient exploration. LLM rewards provide denser semantic signals and modest speedup, but suffer from stochastic variance that introduces noisy gradients. RRM achieves near-Pareto-optimal performance through deterministic, structured reasoning over entropy, throughput-met rate, and energy metrics, enabling faster convergence and higher reward per episode. The GRU+RRM combination is optimal, as \ac{gru}’s temporal modeling fully leverages RRM’s grounded feedback to anticipate scene dynamics and proactively select energy-efficient configurations.

\noindent \textit{(e). Computational efficiency and deployment analysis.}
The GRU-based policy achieves the lowest runtime overhead, requiring only 2.3 ms latency and 45 mJ energy, yielding 82-85\% lower latency and 84-89\% lower energy consumption than ANN, ResNet, and LSTM baselines. This efficiency arises from GRU's lightweight two-gate design, shared temporal weights, and reduced memory transfers through a single hidden state. Furthermore, model and data remapping contribute less than 0.2\% of the total overhead, confirming lightweight execution for adaptive perception on mobile/edge HMPs.

\subsection{Generalization analysis for throughput estimator} 

\begin{figure}
    \centering
    \includegraphics[width=0.99\linewidth]{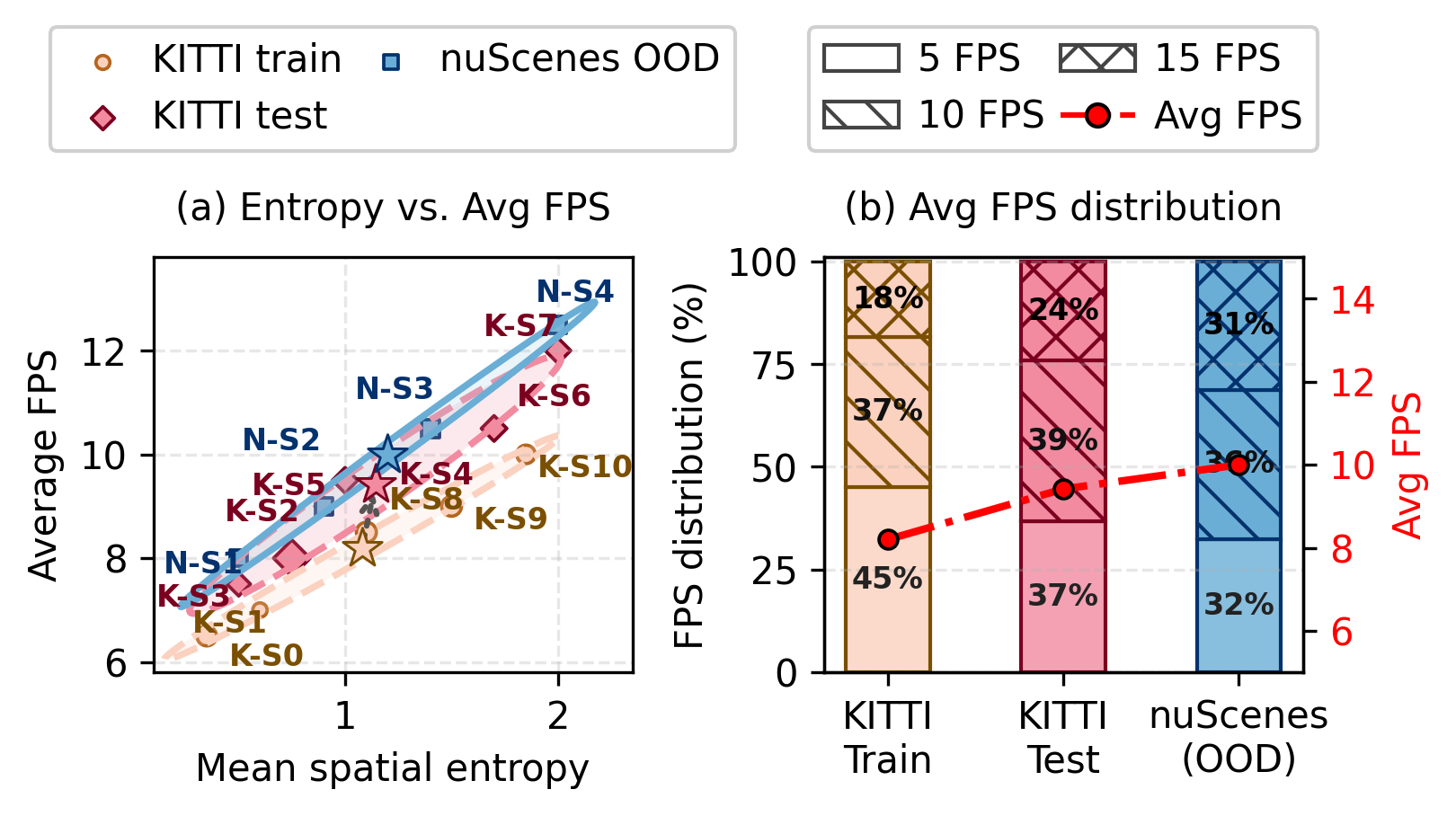}
    \vspace{-18pt}
    \caption{Comparison between KITTI and nuScenes datasets.}
     \vspace{-6pt}
    \label{fig:inter_intra_dataset}
 \end{figure}
 
\begin{figure}
    \centering
    \vspace{-6pt}
    \includegraphics[width=0.99\linewidth]{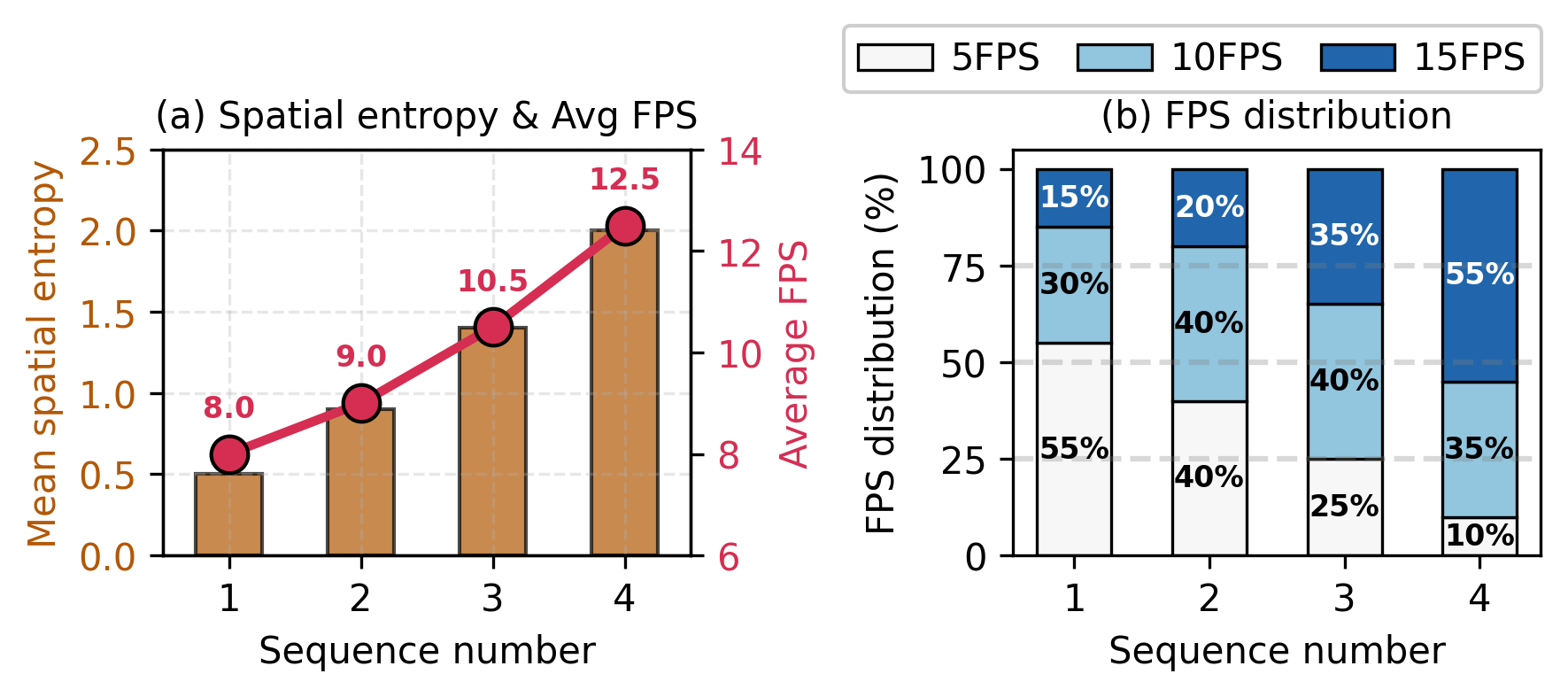}-
    \vspace{-18pt}
    \caption{(a) Impact of mean spatial entropy on throughput (FPS) and (b) throughput (FPS) distribution across the NuScenes dataset's sequences.}
    \vspace{-9pt}
    \label{fig:nuscene_entrpy}
 \end{figure}

\noindent \textit{\textbf{(a).  Zero-shot generalization on Unseen Dataset.}} Figure~\ref{fig:inter_intra_dataset} (a) illustrates both intra and inter-dataset variability in spatial entropy and FPS across \textit{KITTI} (train/test) and nuScenes (out-of-distribution (OOD)) sequences. Within KITTI, the train and test ellipses exhibit partial overlap yet a measurable intra-dataset shift, with test sequences spanning higher entropy (up to 2.0) and achieving a higher average FPS (9.4 vs. 8.2), reflecting greater scene complexity in held-out sequences. Across datasets, the inter-dataset gap is more pronounced: nuScenes sequences cluster at consistently higher entropy and average FPS (10.0), with a 31\% share of 15-FPS frames compared to 24\% for KITTI test and only 18\% for KITTI train, confirming that nuScenes poses a strictly harder OOD scheduling challenge. Figure~\ref{fig:inter_intra_dataset} (b) shows the average FPS distribution across KITTI (train/test) and nuScenes (OOD) sequences. KITTI train is dominated by 5-FPS frames (45\%), reflecting low-complexity scenes where energy savings are readily exploited, whereas nuScenes shifts the balance toward higher FPS, requiring the \textit{TAPAS} to adapt to more frequent, high-complexity frames without prior exposure to this distribution. TAPAS, trained exclusively on KITTI, must generalize to both intra-dataset entropy drift and the broader inter-dataset domain gap, thereby validating its robustness.

\begin{figure}
    \centering
    \vspace{-6pt}
    \includegraphics[width=0.99\linewidth]{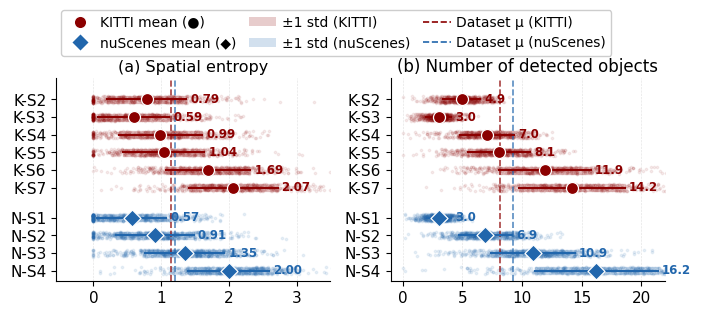}-
    \vspace{-18pt}
    \caption{Sequence-wise spatial entropy and average detected objects distribution for KITTI test sequences and nuScenes OOD sequences.}
    \label{fig:spatial_entpy_eval1}
 \end{figure}

\begin{figure}
    \centering
    \vspace{-6pt}
    \includegraphics[width=0.99\linewidth]{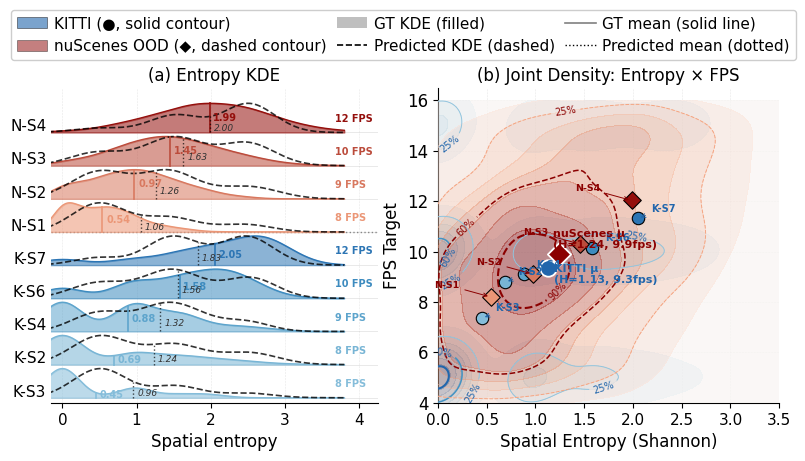}-
    \vspace{-18pt}
    \caption{GT vs. Predicted entropy KDE and cross-dataset joint density.}
    \vspace{-9pt}
    \label{fig:spatial_entpy_eval2}
 \end{figure}

\noindent \textit{\textbf{(b). Spatial entropy: In-Distribution and OOD analysis}}. Figure \ref{fig:spatial_entpy_eval1}(a) and (b) present the per-frame spatial entropy distribution and corresponding detected objects across KITTI test sequences (K-S2-S7) and unseen nuScenes sequences (N-S1-S4), validating the spatial entropy–detected objects mapping. The results show that the entropy preserves a consistent monotonic mapping from detected objects to spatial entropy across both in-dist. and OOD datasets. Figure \ref{fig:spatial_entpy_eval1}(a) shows per-frame entropy scatter, $\pm1$-std bands, and mean values, with dataset-level means indicated by dashed lines ($\mu_{\text{KITTI}}{=}1.13$, $\mu_{\text{nuScenes}}{=}1.22$). KITTI entropy increases monotonically from $\mu_H{=}0.59$ (K-S3) to $\mu_H{=}2.07$ (K-S7), reflecting increasing scene complexity. Low-entropy sequences (K-S2, K-S3) exhibit tight variance, while high-entropy sequences (K-S6, K-S7) show larger intra-sequence variability. nuScenes spans $0.55$--$2.0$, covering and extending the KITTI range: N-S1 and N-S2 lie below the KITTI mean, while N-S3 and N-S4 exceed it, confirming a harder yet non-disjoint OOD distribution. Figure \ref{fig:spatial_entpy_eval1}(b) shows K-S3 exhibits both the lowest entropy ($0.59$) and the fewest detected objects ($3.0$), while K-S7 attains the highest entropy ($2.07$) and the highest detected object count ($14.2$). A consistent monotonic trend is also observed in nuScenes, where N-S1 ($H{=}0.57$, $3.0$ objects) progresses to N-S4 ($H{=}2.00$, $16.2$ objects), with N-S4 exceeding even the densest KITTI sequence in object count, aligning with its higher OOD entropy. Furthermore, it shows that the average number of detected objects across KITTI and nuScenes sequences strongly correlates with spatial entropy, reinforcing its role as a practical proxy for scene-level complexity estimation.

\noindent \textit{\textbf{(c). Robustness analysis of TAPAS on unseen dataset.}}
Figure~\ref{fig:nuscene_entrpy} (a) and (b) present the robustness of the throughput estimator on unseen nuScenes dataset. Figure (a) shows the relationship between scene complexity and the average FPS selected by TAPAS across four nuScenes sequences. As spatial entropy increases from 0.5 (Seq 1) to 2.0 (Seq 4), the average FPS correspondingly rises from approximately 8 FPS to 12 FPS, confirming its ability to match processing rates to the complexity of unseen scenes. Figure \ref{fig:nuscene_entrpy}(b) reveals the FPS distribution shift: Seq 1 (low entropy, simple scenes) operates predominantly at 5 FPS (55\%), whereas Seq 4 (high entropy, complex environments) requires 15 FPS for 55\% of frames. This distribution shift validates our throughput estimator design with a class gap of 5 FPS, which discretizes throughput targets into {5, 10, 15} FPS. This behavior shows that the estimator reliably maps low-complexity frames to low-FPS execution while allocating higher throughput to complex scenes, demonstrating generalization across diverse environments.

\noindent \textit{\textbf{(d). Entropy distribution and joint entropy--FPS density:}}
Figure \ref{fig:spatial_entpy_eval2}(a) and (b) analyzes the throughput estimator using per-sequence entropy KDEs and joint entropy--FPS density distributions across KITTI and unseen nuScenes sequences. Figure \ref{fig:spatial_entpy_eval2}(a) shows predicted entropy KDEs that closely align with the ground-truth entropy distributions for both KITTI and nuScenes. KDEs are used to capture the full entropy distribution, which is not reflected by simple mean and variance. Across both KITTI and unseen nuScenes sequences, the predicted KDEs closely align with the ground-truth distributions, with only minor discrepancies caused by the discrete FPS levels. Figure \ref{fig:spatial_entpy_eval2}(b) presents joint entropy-FPS density contours reveal a strong correlation between scene complexity and assigned FPS targets. While nuScenes is shifted toward higher entropy and FPS values, its distribution partially overlaps with KITTI, confirming a harder yet non-disjoint OOD domain. The consistent entropy--FPS ordering across both datasets validates the estimator's ability to generalize the learned mapping.

 \begin{figure}
    \centering
    \includegraphics[width=0.99\linewidth]{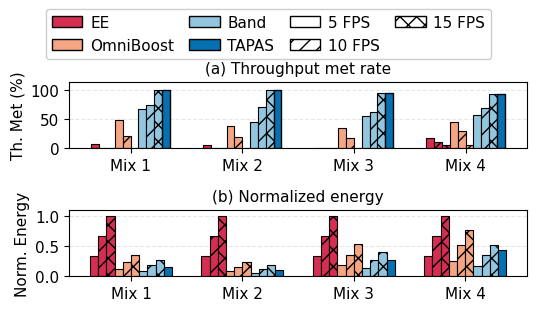}
    \caption{Throughput met (\%) 
     and energy of mix 1-4 with 5, 10, and 15 FPS on nuScenes.}
    \vspace{-9pt}
    \label{fig:th_energy_mixes}
 \end{figure}

\noindent \textit{\textbf{(e). Robustness analysis on unseen dataset: Energy efficiency and throughput met rate}}

For robustness evaluation, we use M1, M2, M3, and M5 mixes with increasing compute intensity using a diverse model pool (see Sec.~\ref{sec.workload}). The mixes progressively increase system complexity. All experiments are conducted on the target embedded platform (see Sec.~\ref{sec.platform}), and we additionally evaluate on the nuScenes dataset to assess the generalization of the throughput estimator on unseen data. We compare TAPAS against SOTA methods (EE~\cite{ee_tro}, OmniBoost~\cite{omniboost}, Band~\cite{band_npu}) across different FPS settings (5, 10, and 15).

Figure~\ref{fig:th_energy_mixes} shows that \textit{TAPAS} achieves 0.10–0.51 normalized energy while maintaining 93–100\% throughput, significantly outperforming static baselines. \textit{EE} incurs the highest energy (1.0 at 15 FPS) and poor throughput ($<$ 52\% at 5 FPS, near 0\% at 15 FPS) due to CPU-only processing. \textit{OmniBoost} reduces energy (0.24–0.77) and improves throughput (68–72\% at 5 FPS), but degrades at higher FPS (13–29\%). \textit{Band} performs best among baselines (0.18–0.77 energy; 62–82\% at 5 FPS), yet fails under dynamic scene variations due to static FPS allocation. In contrast, \textit{TAPAS} leverages throughput adaptive scheduling, assigning 5 FPS to simple scenes and 15 FPS to complex ones, thereby eliminating unnecessary computation. Lower energy in Mix 1–2 (0.10–0.15) arises from efficient GPU usage, while higher energy in Mix 3–4 (0.27–0.51) reflects heavier models. Overall, by predicting throughput requirements and dynamic configurations, \textit{TAPAS} effectively resolves the energy-throughput trade-off. The slight degradation in Mix 3 (95\%) and Mix 4 (93\%) is due to these configurations deploying heavier models (e.g., Swin Transformer, DeepLab v3), where even optimal device selection can occasionally fail to meet tight deadlines during FPS variation.

\subsection{Robustness analysis for hardware unavailability} 

\begin{figure}
    \centering
    \includegraphics[width=0.99\linewidth]{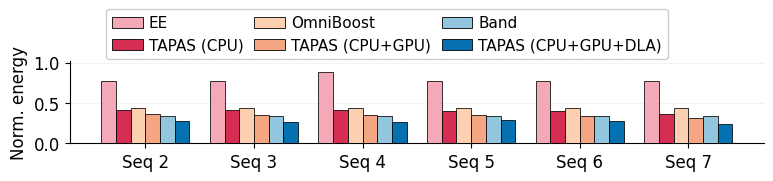}
    \caption{Impact of cluster availability on normalized energy gain}
    \label{fig:cluster_robustness}
    \vspace{-9pt}
\end{figure}

\noindent \textit{\textbf{(a). cluster availability.}}
Figure \ref{fig:cluster_robustness} presents the energy consumption of \textit{TAPAS} across three cluster availability configurations: (i) CPU-only (C1), (ii) CPU+GPU (C2), and (iii) CPU+GPU+DLA (C3), compared against EE, OmniBoost, and Band baselines over KITTI test sequences 2-7. Across all sequences and workload mixes, \textit{TAPAS} consistently achieves the
lowest energy regardless of cluster availability: \textit{TAPAS} (C1-C3)
reduces energy by up to $76\%$, $55\%$, and $35\%$ over EE, OmniBoost,
and Band respectively. This result demonstrates the robustness of \textit{TAPAS} under varying hardware availability. The adaptivity and GRU-based policy adapt effectively even when only CPU resources are available, while larger gains are expected on heterogeneous platforms with GPUs and DLAs due to the increased performance–energy diversity available for task mapping. The scheduling policy explicitly accounts for cluster unavailability during runtime by adapting model-to-cluster mappings based on the currently available CPU, GPU, and DLA resources, ensuring robust performance even under degraded hardware conditions.

\begin{figure}
    \centering
    \includegraphics[width=0.99\linewidth]{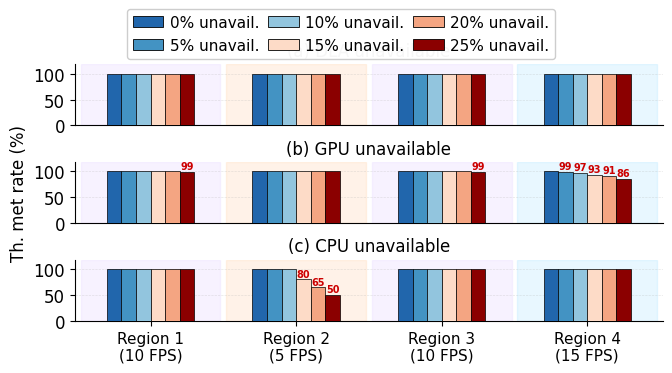}
    \caption{Impact of hardware unavailability on latency compliance of TAPAS. (a). DLA unavailable, (b). GPU unavailable, (c). CPU unavailable. Hardware unavailability is progressively varied from 0-25\% per-cluster. Latency compliance is shown per region in Seq 7}.
    \vspace{-12pt}
    \label{fig:per_clsuter_robustness}
\end{figure}

\noindent \textit{\textbf{(b). Percentage cluster unavailability.}}
Figure \ref{fig:per_clsuter_robustness} evaluates the throughput met rate under progressive cluster unavailability (0-25\%) for seq 7's four region-specific FPS targets (Region 1: 10 FPS, Region 2: 5 FPS, Region 3: 10 FPS, and Region 4: 15 FPS). We achieve conditional cluster preemption by introducing controlled cluster freezing intervals of 50, 100, 150, 200, and 250 ms. This time-multiplexed execution translates into temporal unavailability of clusters, enabling systematic evaluation of robustness under varying cluster unavailability. Across all percentage cluster-unavailable scenarios, \textit{TAPAS} demonstrates strong robustness under progressive cluster unavailability (0-25\%). DLA unavailability has a negligible impact on throughput met rate, while CPU failures affect only specific regions at higher unavailability levels (15--25\%). Throughput met rate remains at 100\% for Regions 1-3 across all GPU unavailability settings, with degradation observed only in the highest-demand case (Region 4, 15 FPS), where performance decreases gracefully from 100\% to 86\% as GPU unavailability increases from 0\% to 25\%. Intermediate GPU unavailability levels (5\%, 10\%, 15\%, and 20\%) result in mild degradation of 1\%, 3\%, 7\%, and 9\%, respectively. In contrast, existing SOTA methods typically assume fixed hardware availability and do not explicitly account for cluster unavailability, making them more prone to significant performance degradation under such conditions. The training traces are collected from real-world execution profiles under varying resource availability conditions, enabling the learned policy to handle 5-15\% cluster unavailability with negligible degradation.

\begin{figure}
    \centering
    \includegraphics[width=0.99\linewidth]{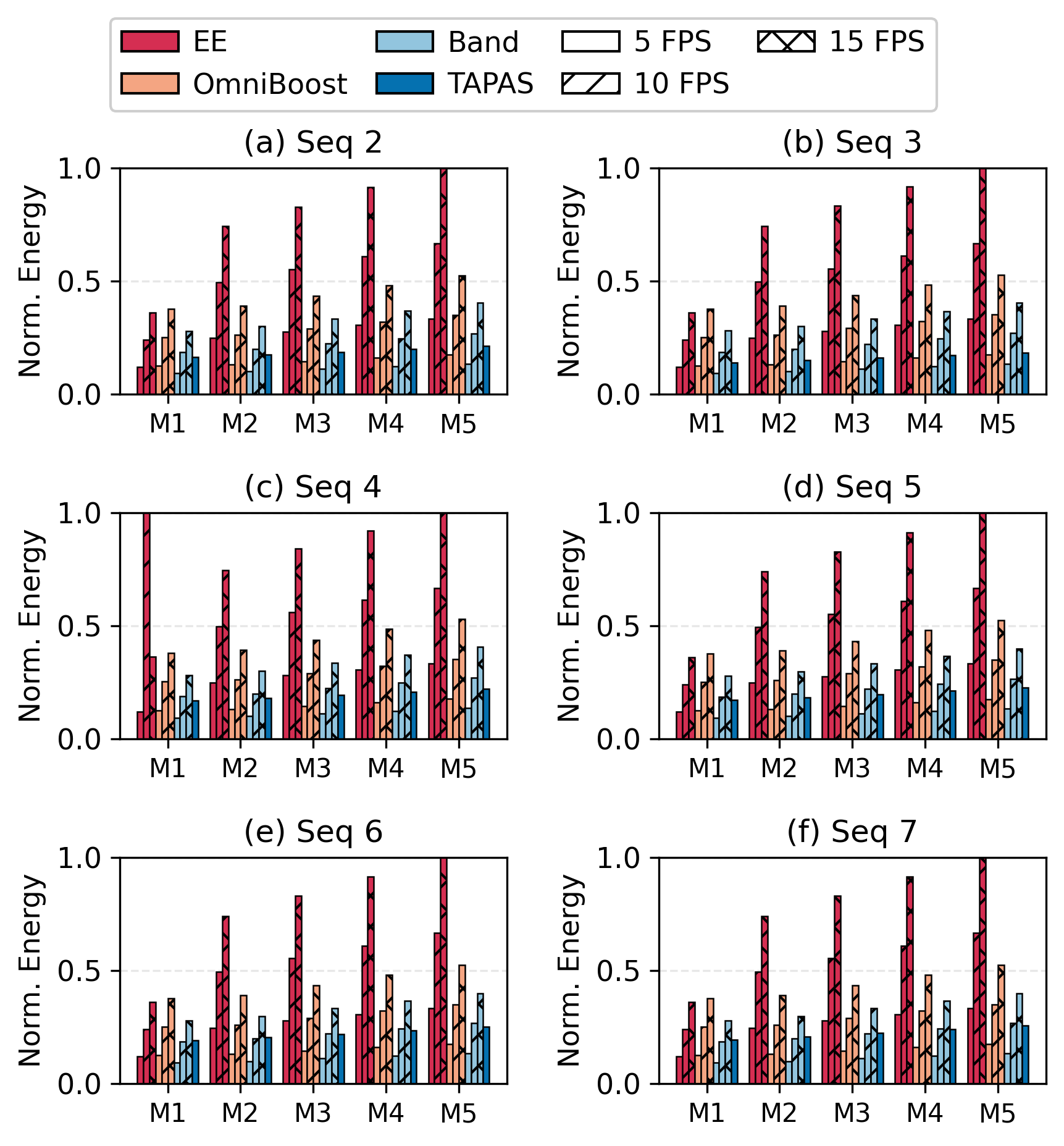}
    \caption{Normalized energy per sequence over different strategies.}
    \label{fig:ene_per_seq}
   \vspace{-9pt}
\end{figure}

\subsection{Static vs adaptivity comparison for SOTA.}

Figure \ref{fig:ene_per_seq} presents normalized energy consumption of six \textit{KITTI} sequences (Seq 2-7) and five different workload mixes (Mix 1-5). For comparison against \textit{TAPAS}, we use three SOTA strategies -- \textit{EE}, \textit{OmniBoost}, \textit{Band} with fixed FPS targets (5, 10, 15 FPS indicated by hatching). \textit{EE} consistently exhibits the highest energy consumption across all sequences and workload mixes. \textit{EE} strategy scales linearly with FPS targets (0.33 at 5 FPS, 0.67 at 10 FPS, 1.0 at 15 FPS), indicating a lack of adaptivity to scene complexity. All SOTA methods process frames at a fixed FPS, regardless of scene complexity, resulting in significantly higher energy consumption. \textit{EE} incurs 79–87\% higher energy than \textit{TAPAS}, while \textit{OmniBoost} achieves moderate savings but still consumes 45–65\% more energy. \textit{Band} performs best among baselines due to sub-graph scheduling, yet fails to adapt to throughput variation across scenes, leaving 35–55\% energy savings unrealized. In contrast, model-level mapping is more effective for transformer execution in our deployment setting, providing better system-level efficiency than subgraph-level scheduling. \textit{TAPAS} achieves the lowest energy by coupling scene awareness with dynamic model-to-cluster mapping, guided by a GRU-based agent with throughput estimation. Its energy scales with scene complexity, delivering up to 87\%, 72\%, and 58\% energy reduction over \textit{EE}, \textit{OmniBoost}, and \textit{Band}, respectively. This shows the advantage of adaptive throughput over fixed FPS with static mapping.

\begin{figure}
    \centering
    \includegraphics[width=0.99\linewidth]{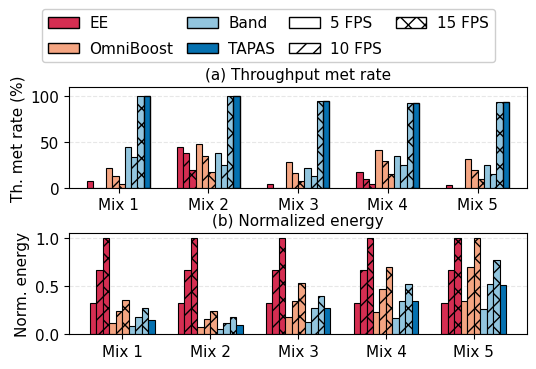}
    \caption{Throughput met (\%) 
     and energy with \textit{EE}, \textit{Omniboost}, \textit{Band}, and \textit{TAPAS} at  5, 10 and 15 FPS with mix-1-5.}
    \label{fig:th_met_mixes}
   \vspace{-12pt}
\end{figure}

Figure~\ref{fig:th_met_mixes} (a) and (b) jointly illustrate normalized energy consumption and throughput met rate for fixed 5/10/15 FPS targets with EE, OmniBoost, and Band, alongside the adaptive TAPAS strategy, across different mix combinations. \textit{EE} exhibits linear energy scaling (2× at 10 FPS, 3× at 15 FPS) due to fixed FPS processing and static mapping, resulting in 69–90\% higher energy than \textit{TAPAS} and poor throughput met rate ($<45$\% at 5 FPS and near 0\% at 15 FPS). \textit{OmniBoost} improves energy (24–77\% vs EE) via CPU–GPU pipelining but achieves only moderate throughput met rate at 5 FPS (22–48\%) and degrades significantly at 15 FPS (5–18\%) due to lack of runtime adaptivity. \textit{Band} leverages accelerators to reduce energy (18–77\% vs EE-OmniBoost) and achieves comparable throughput met rate result to \textit{TAPAS} at 15 FPS (93–100\%) but at a significantly higher energy cost. \textit{Band} fails to adapt to varying demands, resulting in a low throughput-met rate at 5 FPS (22–45\%) and a moderate rate at 10 FPS (63–78\%), thus either missing targets or over-consuming energy. In contrast, \textit{TAPAS} addresses this limitation by dynamically aligning computation with scene-driven FPS targets rather than relying on a fixed FPS with a static mapping. It assigns lower FPS to simple scenes and scales to higher FPS only when the scene demands it, achieving 33–44\% energy savings over \textit{Band} (15 FPS) while maintaining 93–100\% throughput met rate. This is enabled by coupling scene awareness with GRU-based temporal prediction and dynamic model-to-cluster mapping, \ac{rrm}-based multi-objective decision-making, effectively utilizing available hardware only as required, and resolving the energy–throughput trade-off. 

\section{Related work}   
Modern perception pipelines include concurrent tasks of object detection, semantic segmentation, obstacle avoidance, and visual odometry~\cite{tsformer_vo,deepvo}. This creates significant computational and energy demands, making adaptive perception crucial for mobile autonomous systems~\cite{zhuyi_dac,percetion_aware_iccd,suraksha}. We classify existing perception strategies that adapt to scene-awareness and hardware heterogeneity as follows.

\noindent \textbf{Scene-aware adaptivity.}
Existing strategies learn and leverage scene-awareness for adaptive perception. Roborun \cite{roborun_dac} develops spatial-awareness for adaptive perception by prioritizing regions of interest within frames. ADAVP \cite{adavp}, Marlin \cite{marlin}, and SHIFT \cite{context_date} use rate of change in context for scene-awareness, and dynamically select appropriate DNN model settings for adaptivity. In contrast, other context-driven adaptive approaches \cite{adaptive_iros} are primarily simulation-based and lack validation on real heterogeneous edge AI platforms for scene-aware throughput adaptation in \ac{as}. However, aforementioned approaches address only a single vision task and do not consider a comprehensive perception pipeline. Context-based adaptation in these approaches degrades perception quality through model approximation, substituting lightweight or compressed DNN variants when conditions permit, rather than dynamically adjusting throughput targets for existing perception modules. Existing approaches lack mechanisms for estimating throughput, which can adjust FPS based on scene demands

\noindent \textbf{Hardware-aware adaptivity.}
Existing works \cite{ee_iros,ee_tro} allocate the entire perception module to \ac{cpu}, overlooking \ac{hmp}'s low-cost accelerators such as \ac{gpu} and \ac{dla}, and consume significantly more energy. In contrast, the coordinated approaches~\cite{tango,omniboost} extend workload allocation to both CPU and GPU but neglect energy considerations entirely. As design-time solutions, these strategies use fixed allocations that do not adapt to run-time variations in throughput, leading to high energy consumption. Recent works \cite{band_npu, Axonn_dla} consider \ac{npu} and \ac{dla} for resource allocation with basic DNN-specific operator support, while overlooking the inefficiency of transformer inference on \ac{dla}. Although BAND \cite{band_npu} enables finer-grained subgraph-level mapping, such granularity is not effective for transformer workloads in our setting, where system-level efficiency is better achieved through model-level assignment. Existing methods exhibit critical limitations: (1) inability to handle run-time throughput variations \cite{ee_tro,omniboost, band_npu}; (2) low energy efficiency for \ac{as}; (3) improper allocation of perception modules on \ac{hmp} containing low-cost accelerators; \cite{ee_iros,omniboost,tango} and (4) failure to address multi-DNN and transformer scheduling challenges on low-cost accelerators \cite{Axonn_dla,band_npu,tango,ee_tro}. \textit{TAPAS} overcomes these gaps through run-time adaptive resource scheduling for \ac{as}'s perception module, while handling energy-aware run-time variable throughput requirements.

\section{Conclusion} \label{sec.conclusion}
We present \textit{TAPAS}, a throughput-adaptive perception framework for \ac{as} on mobile/edge \acp{hmp}. TAPAS employs spatial entropy for runtime FPS estimation and a GRU-based \ac{rl} agent with \ac{rrm} for optimal task-to-cluster mapping, thereby jointly optimizing throughput met rate and energy consumption. Deployed and evaluated on Jetson Orin NX, TAPAS achieves 93–100\% throughput met rate with up to 76\% energy savings on KITTI sequences, and generalizes to unseen nuScenes with 97\% throughput met rate and 64\% lower energy. Future work will explore model approximation and DVFS for fine-grained energy optimization.






\bibliographystyle{IEEEtran}
\bibliography{references}

\end{document}